\documentclass[journal]{IEEEtran}
\usepackage{amsmath}
\usepackage{algorithm}
\usepackage{algorithmic}
\usepackage{graphicx}
\usepackage{multirow}
\usepackage{tabularx}
\usepackage{xcolor}
\usepackage[table]{xcolor}
\usepackage{array}
\usepackage{amssymb}
\usepackage{bbding}
\usepackage{subfigure}
\usepackage{diagbox}
\usepackage{rotating}
\usepackage{bm}
\usepackage{tikz} 
\usepackage{booktabs}
\usepackage{cite}
\usepackage{CJK}
\usepackage{overpic}
\usepackage[colorlinks =true,
            linkcolor=red,
            anchorcolor=blue,
            citecolor=green
            ]{hyperref}

\newcommand{\addFig}[1]{}
\newcommand{\addFigs}[1]{}

\usepackage{subfigure}
\newcommand{\etal}{\textit{et~al}.~}
\newcommand{\ie}{\textit{i}.\textit{e}.,~}
\newcommand{\eg}{\textit{e}.\textit{g}.,~}

\usepackage{balance}
\usepackage{cleveref}
\crefformat{section}{\S#2#1#3} 
\crefformat{subsection}{\S#2#1#3}
\crefformat{subsubsection}{\S#2#1#3}

\usepackage{colortbl}  
\usepackage{color,xcolor}
\usepackage{soul} 
\soulregister{\cite}7 
\soulregister{\citep}7 
\soulregister{\citet}7 
\soulregister{\ref}7 
\soulregister{\pageref}7 
 
\ifCLASSINFOpdf
\else
\fi

\begin{document}
%
\title{STENet: Superpixel Token Enhancing Network for RGB-D Salient Object Detection}
%
%
%

\author{Jianlin~Chen,
        Gongyang~Li,~\IEEEmembership{Member,~IEEE},
        Zhijiang~Zhang,
        Liang~Chang,
        and~Dan~Zeng,~\IEEEmembership{Senior Member,~IEEE}

\thanks{Jianlin Chen and Zhijiang Zhang are with School of Communication and Information Engineering, Shanghai University, Shanghai 200444, China (email: chen1026@shu.edu.cn; zjzhang@staff.shu.edu.cn).} 
\thanks{Gongyang Li is with the School of Communication and Information Engineering, Shanghai University, Shanghai 200444, China, and Yunnan Key Laboratory of Service Computing, Yunnan University of Finance and Economics, Kunming 650000, China (email: ligongyang@shu.edu.cn).}
\thanks{Liang Chang is with Innovation Academy for Microsatellites of Chinese Academy of Science, Shanghai 201203, China (email: changl@microsate.com).}
\thanks{Dan Zeng is with School of Communication and Information Engineering, Shanghai University, Shanghai 200444, China, and the Institute for Urban Governance, Shanghai University, Shanghai 200444, China (email: dzeng@shu.edu.cn).} 
\thanks{\textit{Corresponding authors: Gongyang Li and Zhijiang Zhang.}}
}

\markboth{}%
{Shell \MakeLowercase{\textit{et al.}}: Bare Demo of IEEEtran.cls for IEEE Journals}

\maketitle

\begin{abstract}
Transformer-based methods for RGB-D Salient Object Detection (SOD) have gained significant interest, owing to the transformer's exceptional capacity to capture long-range pixel dependencies.
Nevertheless, current RGB-D SOD methods face challenges, such as the quadratic complexity of the attention mechanism and the limited local detail extraction.
To overcome these limitations, we propose a novel Superpixel Token Enhancing Network (STENet), which introduces superpixels into cross-modal interaction.
STENet follows the two-stream encoder-decoder structure.
Its cores are two tailored superpixel-driven cross-modal interaction modules, responsible for global and local feature enhancement.
Specifically, we update the superpixel generation method by expanding the neighborhood range of each superpixel, allowing for flexible transformation between pixels and superpixels.
With the updated superpixel generation method, we first propose the Superpixel Attention Global Enhancing Module to model the global pixel-to-superpixel relationship rather than the traditional global pixel-to-pixel relationship, which can capture region-level information and reduce computational complexity.
We also propose the Superpixel  Attention Local Refining Module, which leverages pixel similarity within superpixels to filter out a subset of pixels (\ie local pixels) and then performs feature enhancement on these local pixels, thereby capturing concerned local details.
Furthermore, we fuse the globally and locally enhanced features along with the cross-scale features to achieve comprehensive feature representation.
Experiments on seven RGB-D SOD datasets reveal that our STENet achieves competitive performance compared to state-of-the-art methods.
\end{abstract}


\IEEEpeerreviewmaketitle

\section{Introduction}
\IEEEPARstart{S}{alient}
Object Detection (SOD) stands as a fundamental and crucial task in computer vision. 
Its primary objective is to identify the most visually prominent or important targets within images, providing key prerequisite information for numerous advanced visual processing tasks. 
SOD plays a vital role in industrial defect detection~\cite{li2023no}, target tracking~\cite{LHC2016TIP}, remote sensing scene understanding~\cite{li2023texture,li2023salient}, and medical image analysis~\cite{LGY2021PFOS}. 
However, traditional RGB SOD methods~\cite{zhou2024admnet, zhu2024learning,yun2023towards} often encounter challenges in complex or low-contrast scenes.
To well handle these scenes, researchers have explored the integration of additional information, particularly depth data, and focused on RGB-D SOD.
The combination of depth information has exhibited impressive potential in bolstering the stability and accuracy of RGB-D SOD models in various scenes~\cite{li2020icnet,li2021hierarchical,mou2024salient,he2025samba}. 

Contemporary RGB-D SOD approaches predominantly employ Convolutional Neural Networks (CNNs)~\cite{1989CNN}, with research efforts concentrating on devising multi-modal multi-scale fusion mechanisms~\cite{Wang2019AFNet,Han2018CTMF,Chen2019MMCI}, that effectively harness complementary relationships between heterogeneous visual features. 
Zhang~\etal\cite{zhang2021bilateral} designed a bilateral attention network to learn features from both the foreground and background separately, in order to refine the prediction results. 
However, CNNs exhibit inherent limitations due to their local connectivity, which restricts them to extracting only limited local detail information and hinders their ability to capture global contextual information. 
Their reliance on fixed parameters impedes adaptability across diverse scenarios.

The recent emergence of transformers~\cite{vaswani2017attention} has revolutionized long-range dependency modeling through synergistic integration of self-attention~\cite{yang2022lite} and cross-attention~\cite{lin2022cat} operations, effectively addressing the inherent locality constraints of conventional CNN-based frameworks.
In RGB-D SOD, transformers are typically deployed in dual capacities: serving as backbone networks for hierarchical feature extraction, and functioning as cross-modal integration modules for heterogeneous feature fusion. Wu~\etal\cite{wu2023transformer} effectively captured long-range cross-modal dependencies through parallelized attention mechanisms in the cross-modal interactive parallel transformer module, generating more discriminative fused feature representations. 
However, this fusion method is typically implemented only within the channel dimension due to the quadratic computational complexity issue inherent in transformer attention architectures. 
In order to address this issue, Pang~\etal\cite{pang2023caver} devised a parameter-free operation to implement spatial attention within the transformer, significantly reducing computation and memory requirements. 
However, this fusion method only focuses on the global perspective and pays insufficient attention to local details, making it difficult to accurately capture changes in local details in certain specific scenarios.

Inspired by the above observations, we propose the Superpixel Token Enhancing Network (STENet) based on the two-stream encoder-decoder architecture, which introduces the efficient superpixels to RGB-D SOD.
Superpixels~\cite{shi2000normalized,jampani2018superpixel} aggregate pixels into geometrically coherent units to reduce computational redundancy while preserving object-level structural integrity, which is widely adopted in computer vision.
Here, based on superpixels, we explore global enhancement and local refinement for RGB-D SOD by leveraging superpixels as coherent structural units.
In particular, our STENet contains two specialized superpixel-driven modules for global enhancement and local refinement, named the Superpixel Attention Global Enhancing Module (SAGEM) and the Superpixel Attention Local Refining Module (SALRM), respectively.
SAGEM is responsible for modeling global pixel-to-superpixel relationships instead of traditional pixel-to-pixel interactions, capturing region-level relationships while reducing computational complexity.
SALRM is responsible for identifying and enhancing similar pixels within superpixels, with a focus on fine-grained local details. 
The globally and locally enhanced features generated by SAGEM and SALRM are further fused with the cross-scale features to achieve powerful feature representation.
In this way, the proposed STENet can generate high-quality saliency maps while having efficient computation.

Our main contributions are summarized as follows:
\begin{itemize}
\item We propose a novel Superpixel Token Enhancing Network, which integrates the classical pixel clustering technique, \ie superpixels, into RGB-D SOD to preserve locally coherent structural information while maintaining efficient parallel computing capabilities.

\item We update the superpixel generation method, which expands the neighborhood range of each superpixel, allowing for flexible transformation between pixels and superpixels.

\item We propose the Superpixel Attention Global Enhancing Module to employ a cross-modal superpixel feature for cross-attention mechanisms, resulting in enhanced global feature representation.

\item We propose the Superpixel Attention Local Refining Module to utilize internal statistical information from cross-modal superpixels to implement a self-attention mechanism for selected pixels within the pixel features, achieving enhanced local feature representation.
\end{itemize}

The rest of this paper is organized as follows.
In Sec.~\ref{sec:related}, we review the related work of RGB-D SOD and superpixel representation.
In Sec.~\ref{sec:OurMethod}, we elaborate the proposed STENet.
In Sec.~\ref{sec:exp}, we present comprehensive experiments and analyses.
Finally, in Sec.~\ref{sec:con}, we conclude this work.

\section{Related Work}
\label{sec:related}

\subsection{RGB-D Salient Object Detection}
\label{sec:Tra_ORSI_SOD}

RGB SOD continues to face challenges such as complex backgrounds, low contrast, and varying lighting conditions, which restrict detection accuracy. 
To mitigate these issues, researchers have incorporated depth information into SOD and focused on RGB-D SOD. 
Depth information effectively enhances the ability to distinguish between foreground and background, demonstrating significant application value in complex environments.
In the advancing field of RGB-D SOD, CNN-based methods have achieved remarkable performance due to deep learning advances. 
The existing methods can be categorized into three fusion strategies: early fusion~\cite{liu2019salient,zhao2020single,chen2021rgb}, intermediate fusion~\cite{li2021hierarchical,zhang2021bilateral,fu2021siamese,sun2024bmfnet,ji2021calibrated,sun2021deep,wu2023hidanet,zhang2021depth}, and late fusion~\cite{han2017cnns,wang2019adaptive,chen2020rgbd,pang2020hierarchical}.

Intermediate fusion methods are currently the mainstream scheme, and focus on cross-modal and cross-level interaction and fusion in RGB-D SOD.
For example, Li~\etal\cite{li2021hierarchical} proposed an alternate interaction approach to gradually extract and fuse cross-modal features.
Zhang~\etal\cite{zhang2021bilateral} designed a bilateral attention network to learn features from both the foreground and background separately, in order to refine the prediction results.
Zhang~\etal\cite{zhang2022c} employed a decoupled dynamic filtering convolutional architecture to achieve cross-modal dynamic feature fusion.
Sun~\etal\cite{sun2024bmfnet} presented a dual-branch multi-modal fusion network, aimed at addressing the challenges of multi-modal fusion and multi-level aggregation in RGB-D SOD.
Wu~\etal\cite{wu2023hidanet} enhanced the discrimination ability of RGB and depth features through a granularity-based attention mechanism and introduced a unified cross dual-attention module to achieve multi-modal and multi-level fusion.
However, due to the limitations of convolution, these methods have limited capabilities in modeling global correlations.

Concurrently, the transformer~\cite{vaswani2017attention,liu2021swin}, renowned for its robust capability in modeling long-range dependencies, has garnered increasing attention within the computer vision fields.
In RGB-D SOD, the transformer is frequently employed as backbone network for feature extraction because of its ability to capture rich global context information.
Liu~\etal\cite{liu2021swinnet} utilized the swin transformer~\cite{liu2021swin} to extract features and integrated them with edge information for cross-modal fusion.
Similarly, Liu~\etal\cite{liu2021tritransnet} enhanced multi-level features using three transformer encoders with shared weights.
Hu~\etal\cite{liu2021tritransnet} employed a two-stream swin transformer encoder to extract multi-level and multi-scale features from RGB images and depth images, in order to model global information.

Furthermore, the transformer has been employed for feature interaction and enhancement.
Tang~\etal\cite{tang2022hrtransnet} used inter-feature and intra-feature interactive transformers to aggregate output features of all different resolutions.
Cong~\etal\cite{cong2023point} proposed a cross-modal point-wise interaction module based on transformer attention triggering, exploring feature interaction between different modalities under positional constraints.
Pang~\etal\cite{pang2023caver} proposed a cross-modal perspective-mixed Transformer approach, achieving a balanced trade-off between computational efficiency and accuracy.
However, the transformer-based model also suffers from weaknesses in capturing local information and high computational costs, restricting its application.

\subsection{Superpixel Representation}
\label{sec:SP_SOD}
In the field of computer vision, superpixel representation has long been a focal point of research due to its efficiency as an image representation method.
Traditional superpixel representation techniques mainly rely on unsupervised clustering algorithms, such as graph-based approaches\cite{shi2000normalized}\cite{felzenszwalb2004efficient}, mean shift\cite{comaniciu2002mean}, or K-means clustering\cite{na2010research}\cite{achanta2012slic}.
These methods generate locally consistent pixel blocks (superpixels), which effectively preserve critical structural information within images and facilitate subsequent processing. 

With the emergence of deep learning, the integration of superpixel segmentation with deep learning framework has emerged as a new trend in computer vision.
Varun~\etal\cite{jampani2018superpixel} developed a new differentiable model named SSN for superpixel sampling, utilizing deep networks to learn superpixel segmentation.
Zhang~\etal\cite{zhang2022semantic} combined SLIC with a vision transformer to convert images into superpixel tokens for region-level semantic learning. 
Huang~\etal\cite{huang2023vision} employed sparse connections and local sampling strategies to generate super tokens, effectively reducing computational costs.
Zhu~\etal\cite{zhu2023superpixel} decomposed the pixel space into a low-dimensional superpixel space through local cross-attention for category prediction.
Mei~\etal\cite{mei2024spformer} utilized superpixels to perform irregular yet semantically coherent partitioning of images, effectively capturing intricate image details.
Zhang~\etal\cite{zhang2023lightweight} further integrated superpixels with attention mechanisms in super-resolution, effectively blending global and local attention mechanisms.
 
To address computational redundancy and cross-modal interaction challenges in RGB-D SOD, we introduce a superpixel representation method that compresses RGB-D features into hierarchical superpixel structural units.
Our method achieves efficient cross-modal fusion through dual-phase attention interactions.
Specifically, cross-superpixel attention mechanisms model global contextual relationships, while intensity-geometry guided refinement operations enhance local details within adaptively partitioned superpixel regions.

\begin{figure}
	\centering
	\begin{overpic}[width=1\columnwidth]{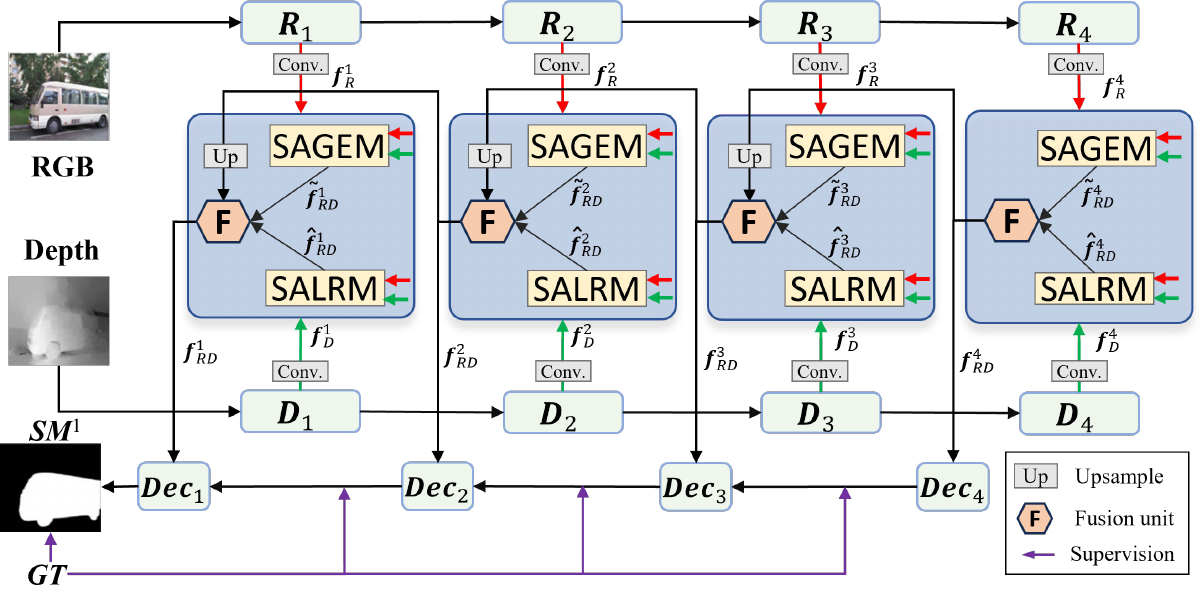}
    \end{overpic}
	\caption{Overview of the proposed STENet. 
    }
    \label{fig:Framework}
\end{figure}

\section{Proposed Method}
\label{sec:OurMethod}
In this section, we present the proposed STENet in detail.
In Sec.~\ref{sec:Overview}, we introduce the network overview.
In Sec.~\ref{sec:CLM}, Sec.~\ref{sec:CAM}, and Sec.~\ref{sec:ESM}, we elaborate the superpixel generation, SAGEM, and SALRM, respectively.
In Sec.~\ref{sec:Loss Function}, we present the loss function.


\subsection{Network Overview}
\label{sec:Overview} 
As depicted in Fig.~\ref{fig:Framework}, our STENet embraces the classic encoder-decoder architecture.
The encoder encompasses RGB and depth branches.
Both branches adopt the Swin-B Transformer~\cite{liu2021swin} for feature extraction with the input sizes of $3\times384\times384$ and $1\times384\times384$, respectively.
Each branch consists of four blocks denoted as $\boldsymbol{R}_{i}/\boldsymbol{D}_{i}$ ($i = 1, 2, 3, 4$), yielding output cross-modal features $f^{i}_{R}/f^{i}_{D}$.
Subsequently, we simultaneously use SAGEM and SALRM to perform global enhancement and local refinement on cross-modal features, generating $\tilde{f}^i_{RD}$ and $\hat{f}_{RD}^i$, respectively.
Then, we adopt the fusion unit to integrate the global feature $\tilde{f}^i_{RD}$, the local feature $\hat{f}_{RD}^i$, and cross-scale features.
In the fusion unit, we first concatenate $\tilde{f}^i_{RD}$ and $\hat{f}_{RD}^i$, then add them to cross-scale features (except for the highest-level one), and finally perform the self-attention operation to capture long-range pixel dependencies.
Finally, our STENet produces the final saliency map $\boldsymbol{SM}^1$ in the decoder.
The decoder consists of four blocks, named $\boldsymbol{Dec}_i$. 
Each block includes a 5$\times$5 depthwise convolution layer, a layer normalization layer, and two 1$\times$1 pointwise convolution layers.
A residual connection adds the output features to the input features, and an upsampling operation adjusts the final sizes.
For comprehensive and efficient supervision, we implement a multi-scale supervision strategy at each block and the final saliency map $\boldsymbol{SM}^1$.
  
\begin{figure}
\centering
  \begin{overpic}[width=1\columnwidth]{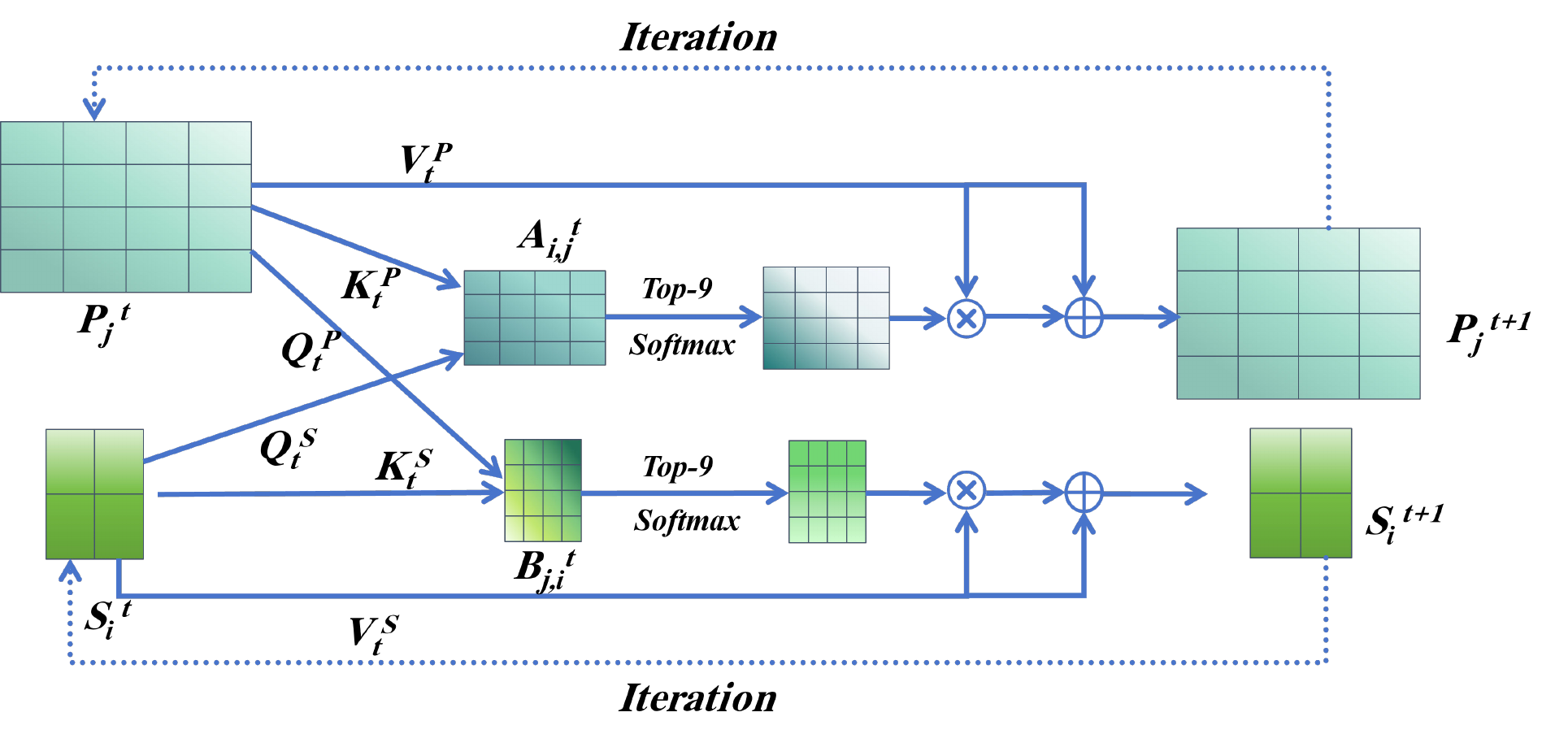}
  \end{overpic}
\caption{
The process of the superpixel generation.
}
\label{SP_TOKEN}
\end{figure}

\subsection{Superpixel Generation}
\label{sec:CLM} 
Distinct from previous methods, we introduce superpixel into RGB-D SOD.
By clustering similar pixels, superpixels preserve crucial local edge information while reducing computational complexity, striking a balance between accuracy and performance.
To enable more flexible and dynamic adjustments of relationships between pixels and superpixels, we adopt the local cross-attention~\cite{zhu2023superpixel} to generate superpixel tokens.
Notably, we expand the neighborhood range of each superpixel.
As shown in Fig.~\ref{SP_TOKEN}, we first initialize the superpixel features \( S\in\mathbb{R}^{M\times d}\) as the average of pixel features \(P\in\mathbb{R}^{N\times d}\), where \(M\) and \(N\) represents the number of superpixels and pixels respectively and \( d \) is the feature dimension.
Then, we obtain the corresponding Query, Key, and Value from both superpixel and pixel features, respectively. 
And we compute the local cross-attention weights \(A_{i,j}^t\) and \(B_{j,i}^t\) between each superpixel and its neighboring pixels, update the superpixel features based on these weights, and repeat this process for \( T \) iteration.

Under normal circumstances, to maintain the locality of superpixels, a masking operation is performed on the acquired weights, ensuring that each pixel is exclusively correlated with the 9 adjacent superpixels. 
However, in shallow layers, high-resolution features lead to dense superpixel grids (\eg 96$\times$96 pixels), where 3$\times$3 neighborhoods may miss semantically relevant superpixels due to strict spatial constraints. Expanding to 5$\times$5 allows pixels to search for similar superpixels over a wider spatial range, improving the quality of selected features.


For each pixel \(i\), we choose the top-9 most similar superpixels from its correlation map \(A_{i,j}^t\) and set the corresponding elements in \(\Omega_{9,i,j}^t\) to 1, with the rest set to 0.
Similarly, for each superpixel \(j\), we select the top 9\( \times(M/N)\) most similar pixels from its correlation map \(B_{j,i}^t\) and set \(\Omega_{9,j,i}^t\) to 1, with the others set to 0.
By maintaining the top-9 selection (same as 3×3 neighborhoods), the computational complexity remains constant. The mask operation still filters only the most relevant superpixels, avoiding redundant calculations while leveraging the expanded spatial context.
Using these masks, we update the features as follows:
\begin{equation} 
  \begin{aligned}
  {S}_i^{t+1} = \text{softmax}( \sum_{j} \Omega_{k,i,j}^t A_{i,j}^t ) \odot V_{t,i}^S + S_i^{t},\\
  {P}_j^{t+1} = \text{softmax}( \sum_{i} \Omega_{h,j,i}^t B_{j,i}^t ) \odot V_{t,j}^P + P_j^{t},    
  \end{aligned}
  \label{eq:update_pixel}  
\end{equation}  
where \(\odot\) denotes the element-wise multiplication, and \(S_i^{t}\) and \(P_j^{t}\) represent the superpixel and pixel features from the previous iteration, respectively.
This approach combines the retention of essential information with residual learning.
Through $T$ iterations of this process, we ultimately get the refined superpixel features ${S}_i^T$. In practical implementation, $T$ is set to 2. This is because excessive iterations offer no benefit to performance but significantly increase computational costs. 
\begin{figure}[t!]
\centering
  \begin{overpic}[width=1\linewidth]{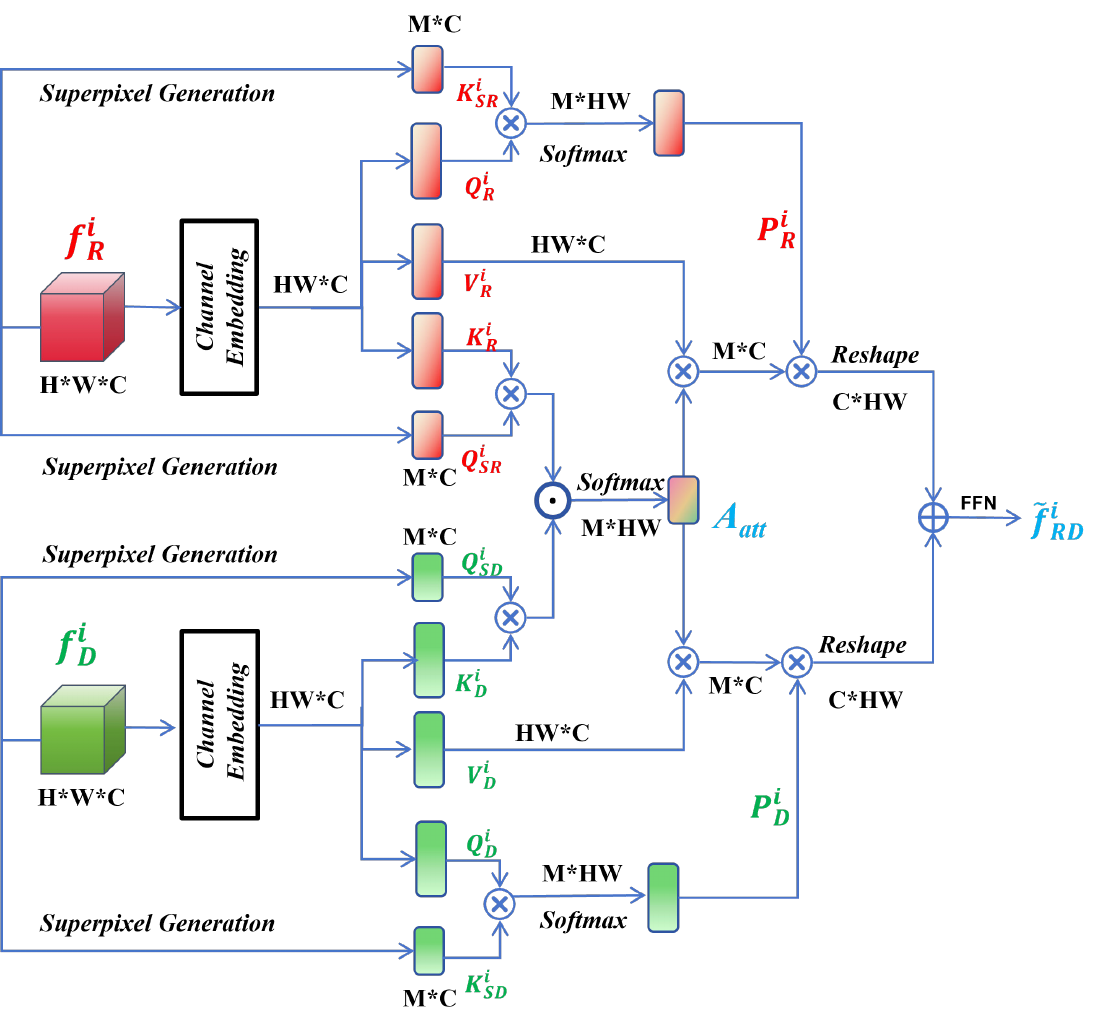}
  \end{overpic}
\caption{
Illustration of the Superpixel Attention Global Enhancing Module.
}
\label{CAM_structure}
\end{figure}
\subsection{Superpixel Attention Global Enhancing Module}
\label{sec:CAM} 
The use of transformers for cross-modal feature interaction and enhancement of global information has become widespread. 
However, in most cases, only the channel dimension is considered due to the heavy computational workload in the spatial dimension. 
Therefore, we propose a Superpixel Attention Global Enhancement Module (SAGEM), which establishes a relationship between superpixels and pixels in the spatial dimension, thereby facilitating cross-modal feature interaction and enhancing the dissemination of global information.

As depicted in Fig.~\ref{CAM_structure}, the inputs of SAGEM are $f_{R}^{i}$ and $f_{D}^{i}$.
We flatten their sizes from \(H \times W \times C\) to \(HW \times C\) and perform linear embedding to acquire the respective query ($Q^{i}_{R}/Q^{i}_{D}$), key ($K^{i}_{R}/K^{i}_{D}$), and value ($V^{i}_{R}/V^{i}_{D}$). 
We begin by applying the superpixel generation method, as detailed in Sec.~\ref{sec:CLM}, to obtain superpixels \(Q^i_{SR}\) and \(Q^i_{SD}\) for each modality. 
These superpixels are subsequently flattened to a dimension of $M \times C$, where \(M\) represents the quantity of superpixels and \(C\) denotes the number of channels.

Then, we establish dependencies between pixels and superpixels via the local cross-attention mechanism, and then utilize superpixels for global information propagation.
We compute the global attention maps \(A^i_{R}\) and \(A^i_{D}\) through matrix multiplication between the queries of superpixels and the keys of pixels, yielding attention maps with sizes of $M\times HW$ as follows:
\begin{equation}
  \begin{aligned}
      A_{R}^i = \text{softmax}\left(\frac{{Q^i_{SR}} \otimes {K^i_{R}}^{\top}}  {\sqrt{D_{k}}}\right),\\
      \quad A^i_{D} = \text{softmax}\left(\frac{{Q^i_{SD}} \otimes {K^i_{D}}^{\top}}  {\sqrt{D_{k}}}\right),
      \label{eq:2}
   \end{aligned}
\end{equation}
where $\otimes$ represents the matrix multiplication.
This process facilitates the derivation of attention maps with sizes of $M \times HW$, where \(M\) represents the number of superpixels and \(HW\) corresponds to the flattened pixel dimensions.
The softmax function serves to normalize the values, while \(D_{k}\) operates as a scaling factor to mitigate gradient vanishing.
To facilitate information interaction among cross-modal features, we perform an element-wise multiplication between the global attention maps of the two modalities, yielding a shared attention map $A_{att}^i$:
\begin{equation}
A_{att}^i = A^i_{R} \odot A^i_{D}.
 \label{eq:3}
\end{equation}

We then multiply the shared attention map \(A_{att}\) with the corresponding value matrices, generating outputs of dimensions $M \times C$. 
Global features of the two modalities interact through superpixels, and the co-attentioned features are enhanced.
Next, we employ another self-attention operation for the two modalities. By utilizing superpixel keys and pixel queries, we distribute the global information from the superpixels into the pixel space, obtaining attention maps \(P^i_{R}\) and \(P^i_{D}\) as follows:
\begin{equation}
  \begin{aligned}
    P^i_{R} &= \textit{Softmax}\left(\frac{K^i_{SR} \otimes {Q^i_{SR}}^{\top}}{\sqrt{D_{k}}}\right), \\
    P^i_{D} &= \textit{Softmax}\left(\frac{K^i_{SD} \otimes {Q^i_{SD}}^{\top}}{\sqrt{D_{k}}}\right).
    \label{eq:4}
  \end{aligned}
\end{equation}
These mapping matrices are utilized to propagate the updated superpixel feature information back into the original pixel space:
\begin{equation}
    \tilde{f}^i_{RD} = P^i_{R} \otimes (A^i_{att} \otimes V^i_{R})+ P^i_{D} \otimes (A^i_{att} \otimes V^i_{D}),
   \label{eq:5}
\end{equation}
where \(\tilde{f}^i_{RD}\) signifies the enhanced features remapped to the original pixel space.

\begin{figure}
	\centering
	\begin{overpic}[width=1\columnwidth]{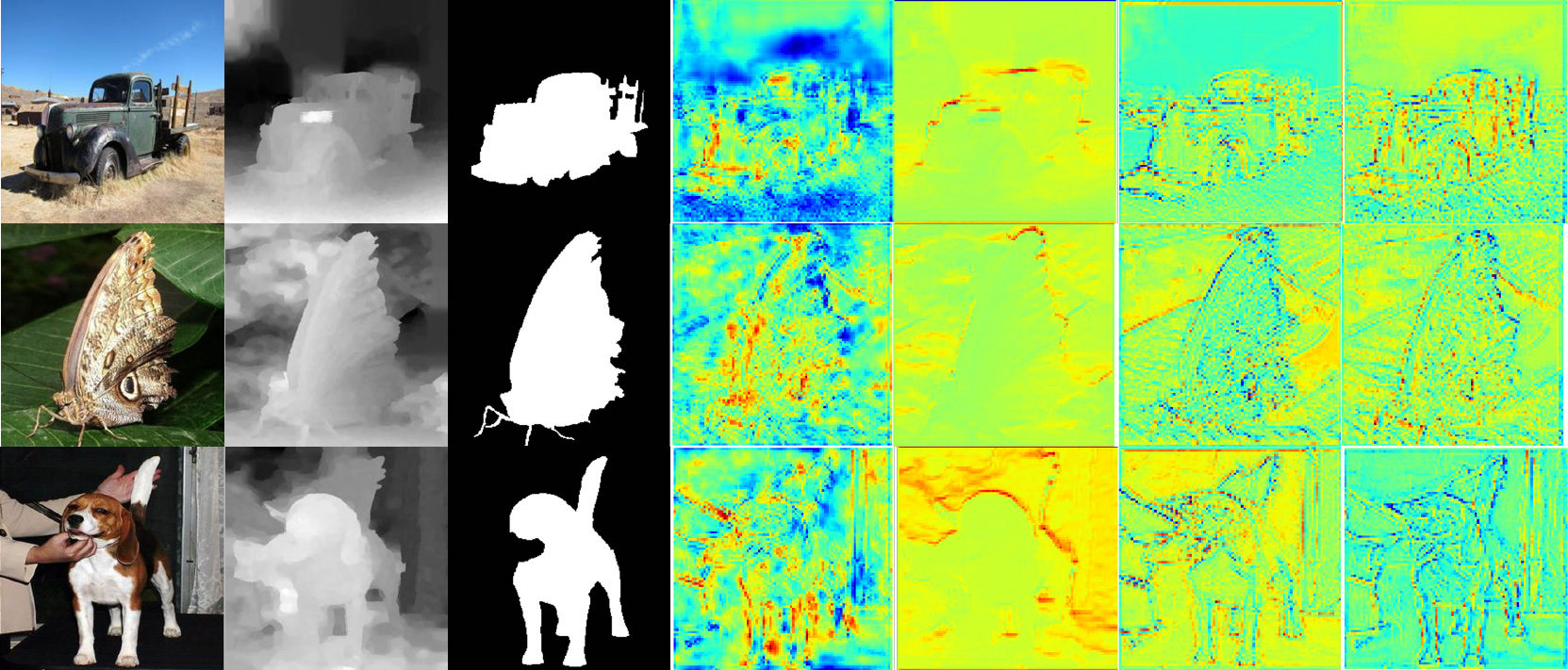}

        \put(5, -4){(a) }     
		\put(19, -4){(b) }   
		\put(34, -4){(c) }
        \put(47, -4){(d)}
		\put(62, -4){(e)}
        \put(77, -4){(f)}
        \put(89, -4){(g)}
    
    \end{overpic}
	\caption{Visualization of features in SAGEM and SALRM. (a) RGB, (b) depth, (c) GT, (d) RGB features, (e) depth features, (f) SAGEM features, and (g) SALRM features.} 
    
    \label{fig:Visualization}
\end{figure}
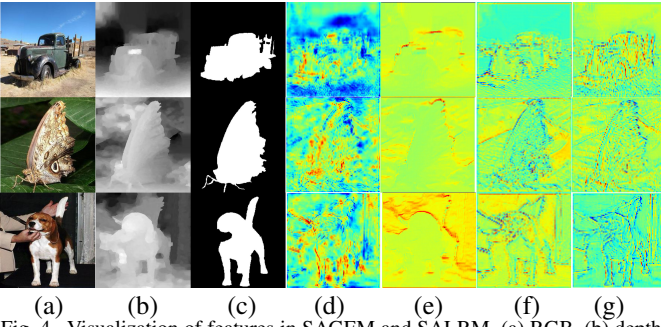

Finally, the globally enhanced features \(\tilde{f}^i_{RD}\) are further transformed and output through a Feedforward Network (FFN).
We visualize features in our SAGEM in Fig. \ref{fig:Visualization} (f).
The SAGEM extracts features that prioritize the integration of cross-modal global contextual information. By capturing long-range dependencies, the module enhances the overall contours and semantic consistency of objects.
The attention mechanism design enables our model to focus on cross-regional semantic associations. For instance, it can connect scattered object components into a coherent structure via contextual cues, thereby improving the overall performance of objects in complex scenes.

\subsection{Superpixel Attention Local Refining Module}
\label{sec:ESM}
%
%
Although SAGEM can enhance global features, it inevitably loses some local information.
Additionally, superpixel token generation may not be precise due to noise and other factors, resulting in semantic confusion of different regions.
To address the above issues,
we propose a Superpixel Attention Local Refining Module (SALRM) to utilize pixel similarity within superpixels to significantly enhance local features, boosting the model's ability to capture and express details.

\begin{figure}[t!]
   \centering
    \begin{overpic}[width=1\linewidth]{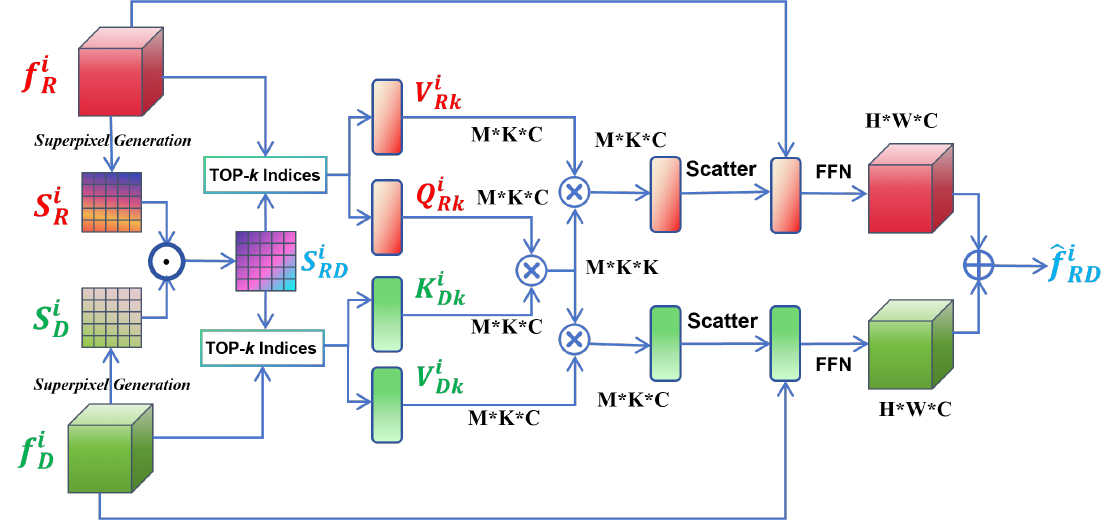}
    \end{overpic}
    \caption{
Illustration of the Superpixel Attention Local Refining Module.
}  
\label{ESM_structure}
\end{figure}

As shown in Fig.~\ref{ESM_structure}, we first perform superpixel generation on the input RGB and depth features of SALRM, generating association matrices, denoted as $S_R^i$ and $S_D^i$.
To address the potential semantic confusion caused by superpixel generation, we need to filter the co-associated pixels of superpixels to highlight local features.
The association matrix represents the weight relationship between each pixel and superpixels.
By performing element-wise multiplication on the association matrices of the two modalities, we generate a combined association matrix $S_{RD}^i$ to determine the pixels that are semantically best-matched to the superpixel tokens:
\begin{equation}
  S_{RD}^i = S_R^i \odot S_D^i .
  \label{eq:6}
\end{equation}

For the combined association matrix, the weights of pixels that are similar within superpixels of both modalities are enhanced.
Next, we need to determine which pixel features require individual local enhancement by sorting the weights.
Given a superpixel denoted as $f = \{x(i)\}_N \in \mathbb{R}^{N \times C}$, encompassing numerous pixels where $x(i)$ represents the pixel, $N$ signifies the count of selected pixels and $C$ represents the channel count).
It is worth noting that the pixel count within each superpixel can fluctuate. 
To streamline computations, we exclusively amplify the $k$ most similar pixels in each superpixel. 
Through the application of the top-$k$ function, we efficiently pinpoint the indices of these remarkably similar pixels.

Subsequently, we perform linear embedding on the input features to generate corresponding queries, keys, and values. Utilizing the previously acquired index data, we employ the gather function to extract values from $Q$, $K$, and $V$ that are aligned with each superpixel position, and then appropriately rearrange their dimensions.
The resultant local feature set possesses a dimension of $M\times N\times C$, where $M$ indicates the superpixel count and $N$ signifies the number of extracted components.

For these pixels with similar semantic information, we utilize the cross-modal local features $Q_{Rk}^i$ and $K_{Dk}^i$ to construct a local attention map through the cross-attention mechanism, thereby facilitating cross-modal information interaction of local features.
The precise computation procedure is outlined below:
\begin{equation}
  F_{att} = \text{Softmax}\left(\frac{{Q}^i_{Rk}  \otimes {{K}^i_{Dk}}^{\top} } {\sqrt{D_k}}\right). 
  \label{eq:7}
\end{equation}
Following this, we execute the matrix multiplication between the generated local attention map and the respective local values $V_R^i$ and $V_D^i$, thereby obtaining the refined local features:
\begin{equation}
  \begin{aligned}
    \hat{f}_{Rk}^i &= F_{att} \otimes V_{Rk}^i, \\
    \hat{f}_{Dk}^i &= F_{att} \otimes V_{Dk}^i.
  \end{aligned}
  \label{eq:8}
\end{equation}

Utilizing the previously secured index value information, we scatter the enhanced local features back to their original positions and integrate them with the original features.
We then employ the FFN to further refine the feature expression.
Ultimately, we perform the element-wise summation (\ie $+$) to merge the enhanced features from both modalities, deriving the ultimate feature representation $\hat{f}_{RD}^i$:
\begin{equation}
  \hat{f}_{RD}^i = \text{FFN}(\hat{f}_{Rk}^i + \hat{f}_R^i) + \text{FFN}(\hat{f}_{Dk}^i + \hat{f}_D^i).
  \label{eq:9}
\end{equation}

By maximizing the utilization of the information embedded within the association matrices generated by superpixels, we can precisely pinpoint the most similar pixels within a superpixel and harness a cross-attention mechanism for local feature refinement. 
We visualize features in our SALRM in Fig. \ref{fig:Visualization} (g).
The SALRM extracts features that prioritize enhancing local structural details (\eg textures and edges). The attention mechanism suppresses noise by strengthening intra-superpixel feature similarity, refines object boundaries, and enhances the structural representation accuracy of salient regions.

\subsection{Loss Function}
\label{sec:Loss Function}
We utilize a hybrid loss function that combines Binary Cross-Entropy (BCE) loss and Intersection over Union (IoU) loss for comprehensive supervision.
This hybrid approach ensures meticulous supervision at both pixel level and patch level. 
Furthermore, we incorporate a deep supervision strategy to bolster the network's supervision power and expedite convergence.

The overall loss function $L_{\text{total}}$ is computed as:
\begin{equation}  
    \begin{aligned}  
L_{\text{total}} &= \underbrace{-\frac{1}{N} \sum_{i=1}^{N} \left[ g_i \log(s_i) + (1 - g_i) \log(1 - s_i) \right]}_{\text{BCE Loss}} \\  
&+ \underbrace{1 - \frac{\sum_{i=1}^{N} s_i \times g_i}{\sum_{i=1}^{N} \left( s_i + g_i - s_i \times g_i \right)}}_{\text{IoU Loss}} , 
    \end{aligned}  
\end{equation}
where $N$ denotes the total number of pixels, $s_i$ signifies the saliency value for the $i$-th pixel in the predicted saliency map, and $g_i$ corresponds to the ground truth value for the same $i$-th pixel in $\boldsymbol{GT}$.

\section{Experiments}
\label{sec:exp}

\begin{table*}[t!]
  \centering
 \renewcommand{\arraystretch}{1.0}
 \renewcommand{\tabcolsep}{0.4mm}
  \caption{
  Quantitative comparisons (\%) on the test set of NJUD\cite{ju2014depth}, NLPR\cite{peng2014rgbd}, and DUT-RGBD\cite{piao2019depth} dataset.
  The top two results in each column are highlighted in \textcolor{red}{\textbf{red}} and \textcolor{blue}{\textbf{blue}}.
    }
\label{table:QuantitativeResults1}
 \begin{tabular}{r|ccc|cc|*{5}{>{\centering\arraybackslash}p{2em}}|*{5}{>{\centering\arraybackslash}p{2em}}|*{5}{>{\centering\arraybackslash}p{2em}}} 
    \toprule
    \multirow{2}{*}{{Model}} & 
    \multirow{2}{*}{{Publisher}}&
    \multirow{2}{*}{{Year}}&
    \multirow{2}{*}{{Backbone}}&
     \#Param &
     FLOPs &
    \multicolumn{5}{c|}{{NJUD\cite{ju2014depth}}} & \multicolumn{5}{c|}{{NLPR\cite{peng2014rgbd}}} & \multicolumn{5}{c}{{DUT-RGBD\cite{piao2019depth}}}\\  

    \cline{7-21}
     &  &  &  & (M)↓ & (G)↓ & \multicolumn{1}{c}{$M$↓} & \multicolumn{1}{c}{$F_{\beta}$↑} & \multicolumn{1}{c}{$E_{m}$↑} & \multicolumn{1}{c}{$S_{m}$↑} & \multicolumn{1}{c|}{$F^{\omega}_{\beta}$↑} & \multicolumn{1}{c}{$M$↓} & \multicolumn{1}{c}{$F_{\beta}$↑} & \multicolumn{1}{c}{$E_{m}$↑} & \multicolumn{1}{c}{$S_{m}$↑} & \multicolumn{1}{c|}{$F^{\omega}_{\beta}$↑} & \multicolumn{1}{c}{$M$↓} & \multicolumn{1}{c}{$F_{\beta}$↑} & \multicolumn{1}{c}{$E_{m}$↑} & \multicolumn{1}{c}{$S_{m}$↑} & \multicolumn{1}{c}{$F^{\omega}_{\beta}$↑} \\  
    \midrule
   
    DSNet~\cite{wen2021dynamic} & TIP & 2021  & Res50 & - & -  & .034 & .922 & .947 & .921 & .898 & .024 & .917 & .956 & .926 & .886 & .079 & .821 & .886 & .841 & .774  \\
    DCFNet~\cite{ji2021calibrated} & CVPR & 2021  & Res50 & 108.5 & 107.8 & .035 & .915 & .944 & .911 & .893 & \textcolor{blue}{.021} & .912 & .959 & .923 & .892 & .071 & .835 & .891 & .836 & .766  \\
    CIRNet~\cite{cong2022cir} & TIP & 2022 & Res50 & 103.1 & 22.5 & .035 & .927 & .943 & .925 & .895 & .023 & .924 & .955 & \textcolor{blue}{.933} & .889 & .031 & .940  & .958 & .932 & .904  \\
    C2DFNet~\cite{zhang2022c} & TMM & 2022  & Res50 & 47.5 & 22.1 & .038 & .909 & .941 & .907 & .885 & \textcolor{blue}{.021} & .916 & .961 & .927 & .897 & \textcolor{red}{ \textbf .026} & \textcolor{blue}{.943} & \textcolor{blue}{\textbf .963} & .933 & \textcolor{blue}{\textbf .918} \\
    DIGRNet~\cite{cheng2023depth} & TMM & 2022  & Res50 & 201.8 &68.2 & \textcolor{red}{\textbf.028} & \textcolor{red}{ \textbf.936} & \textcolor{red}{ \textbf.954} & \textcolor{red}{ \textbf.933} & \textcolor{blue}{.909} & .023 & \textcolor{blue}{.928} & .956 & \textcolor{red}{ \textbf.935} & .895 & .033 & .938 & .951 & .926 & .898  \\
    CFIDNet~\cite{chen2022cfidnet} & NCA & 2022  & Res50 & 53.9 &42.9 & .038 & .914 & .929 & .914 & .886 & .026 & .906 & .951 & .922 & .881 & .039 & .920  & .944 & .916 & .887  \\
    HINet~\cite{bi2023cross} & PR & 2023  & Res50 & 98.9 & 389.7& .039 & .913 & .938 & .915 & .881 & .026 & .906 & .950  & .922 & .876 & .054 & .873 & .921 & .884 & .826  \\
    CAVER~\cite{pang2023caver} & TIP & 2023  & Res50 & 55.5 & 44.3 & .031 & .924 & .953 & .920  & .906 & \textcolor{red}{ \textbf.020} & .921 & \textcolor{blue}{.963} & .928 & \textcolor{blue}{.901} & .042 & .904 & .937 & .903 & .874  \\
    RD3D+~\cite{chen20223} & TNNLS & 2024  & Res50 & 28.9 &43.3 & .033 & .928 & .948 & \textcolor{blue}{.927} & .899 & .022 & .920 & .958 & \textcolor{blue}{.933} & .889 & .031 & \textcolor{red}{ \textbf.944} & .960 & \textcolor{red}{ \textbf.936} & .908  \\
    LAFB~\cite{wang2024learning}  & TCSVT & 2024 & Res50 & 451.8 & 137.6 & .033 & .918 & .947 & .916 & .897 & .027 & .909 & .946 & .913 & .876 & \textcolor{blue}{.030} & .935 & .959 & .927 & .909  \\
    \textbf{Ours} &       & 2025  & Res50 & 56.4 & 25.5 & \textcolor{blue}{.029} & \textcolor{blue}{.930} & \textcolor{blue}{.953} & .923 & \textcolor{red}{ \textbf.912} & \textcolor{red}{ \textbf.020} & \textcolor{red}{ \textbf.929} & \textcolor{red}{ \textbf.964} & \textcolor{blue}{.933} & \textcolor{red}{ \textbf.909} & \textcolor{red}{ \textbf.026} & \textcolor{red}{ \textbf.944} & \textcolor{red}{ \textbf.966} & \textcolor{blue}{.935} & \textcolor{red}{ \textbf.922}  \\
    \hline
    \hline
    SwinNet~\cite{liu2021swinnet} & TCSVT & 2022  & SwinB & 198.7 &124.3 & .027 & \textcolor{blue}{.938} & .954 & \textcolor{blue}{.934} & .917 & .018 & \multicolumn{1}{c}{\textcolor{blue}{.936}} & \multicolumn{1}{c}{.969} & \textcolor{red}{.941} & \multicolumn{1}{c|}{.913} & .021 & .958 & .973 & .948 & .933 \\
    CATNet~\cite{sun2023catnet} & TMM & 2023  & SwinB & 262.6 & 341.8 & .026 & .937 & .957 & .932 & .922 & .018 & .934 & \multicolumn{1}{c}{.970} & \textcolor{blue}{.940} & \multicolumn{1}{c|}{.916} & .020 &\textcolor{blue}{ .960} & .974 & \textcolor{red}{ \textbf.953} & .942  \\
    PICRNet~\cite{cong2023point} & ACM MM & 2023  & SwinB &106.8 & 121.3 & .029 & .930  & .950  & .927 & .912 & .019 & .931 & .967 & .935 & .911 & .021 & .953 & .970 & .943 & .933  \\
    CPNet~\cite{hu2024cross} & IJCV & 2024  & SwinB & 216.5 & 129.3 & \textcolor{blue}{.025} & \textcolor{red}{.941} & \textcolor{blue}{ \textbf.960} & \textcolor{red}{.935} & \textcolor{blue}{ \textbf.927} & \textcolor{blue}{.016} & \multicolumn{1}{c}{\textcolor{blue}{.936}} & \multicolumn{1}{c}{\textcolor{blue}{.972}} & \textcolor{blue}{.940} & \multicolumn{1}{c|}{\textcolor{blue}{.922}} & \textcolor{blue}{.019} & \textcolor{red}{.961} & \textcolor{blue}{.976} & \textcolor{blue}{.951} & \textcolor{blue}{.948}  \\
    \textbf{Ours} &   TMM    & 2025  & SwinB & 180.5 & 118.3 & \textcolor{red}{ \textbf.023} & \textcolor{red}{ \textbf.941} & \textcolor{red}{.967} &  .932 & \textcolor{red}{.931} & \textcolor{red}{ \textbf.015} & \textcolor{red}{ \textbf.938} & \textcolor{red}{ \textbf.975} & .939 & \textcolor{red}{ \textbf.929} & \textcolor{red}{ \textbf.018} & \textcolor{red}{ \textbf.961} & \textcolor{red}{ \textbf.979} & .950 & \textcolor{red}{ \textbf.952} \\
    \bottomrule
    \end{tabular}%

\end{table*}

\begin{table*}[t!]
   
   \renewcommand{\arraystretch}{1.}
   \renewcommand{\tabcolsep}{0.5mm}
   \centering 
    \caption{
  Quantitative comparisons (\%) on the test set of LFSD\cite{li2014saliency}, SIP\cite{fan2020rethinking}, STEREO\cite{niu2012leveraging}, and SSD\cite{zhu2017three} dataset.
  The top two results in each column are highlighted in \textcolor{red}{\textbf{red}} and \textcolor{blue}{\textbf{blue}}.
    }
\label{table:QuantitativeResults2}
 
     \begin{tabular}{r|c|*{5}{>{\centering\arraybackslash}p{2em}}|*{5}{>{\centering\arraybackslash}p{2em}}|*{5}{>{\centering\arraybackslash}p{2em}}|*{5}{>{\centering\arraybackslash}p{2em}}} 
    \toprule
    \multirow{2}{*}{{Model}} &
    \multirow{2}{*}{{Backbone}} & \multicolumn{5}{c|}{{LFSD\cite{li2014saliency}}}& \multicolumn{5}{c|}{{SIP\cite{fan2020rethinking}}} & \multicolumn{5}{c|}{{STEREO\cite{niu2012leveraging}}} & \multicolumn{5}{c}{{SSD\cite{zhu2017three}}}\\  
    \cline{3-22}
    & & \multicolumn{1}{c}{$M$↓} & \multicolumn{1}{c}{$F_{m}$↑} & \multicolumn{1}{c}{$E_{m}$↑} & \multicolumn{1}{c}{$S_{m}$↑} & \multicolumn{1}{c|}{$F^{\omega}_m$↑} & \multicolumn{1}{c}{$M$↓} & \multicolumn{1}{c}{$F_{m}$↑} & \multicolumn{1}{c}{$E_{m}$↑} & \multicolumn{1}{c}{$S_{m}$↑} & \multicolumn{1}{c|}{$F^{\omega}_m$↑} & \multicolumn{1}{c}{$M$↓} & \multicolumn{1}{c}{$F_{m}$↑} & \multicolumn{1}{c}{$E_{m}$↑} & \multicolumn{1}{c}{$S_{m}$↑} & \multicolumn{1}{c|}{$F^{\omega}_m$↑} & \multicolumn{1}{c}{$M$↓} & \multicolumn{1}{c}{$F_{m}$↑} & \multicolumn{1}{c}{$E_{m}$↑} & \multicolumn{1}{c}{$S_{m}$↑} & \multicolumn{1}{c}{$F^{\omega}_m$↑} \\  
    \midrule
   
   DSNet~\cite{wen2021dynamic} & Res50 & .069 & .858 & .905 & .868 & .826 & .052 & .881 & .920 & .876 & .840 & .036 & .911 & \textcolor{blue}{.947} & \textcolor{blue}{.915} & .882 & .045 & \textcolor{red}{ \textbf.878} & \textcolor{blue}{.923} & \textcolor{red}{ \textbf.885} & \textcolor{blue}{.838} \\
    DCFNet~\cite{ji2021calibrated} & Res50 & .075 & .841 & .883 & .841 & .805 & .051 & .884 & .921 & .875 & .848 & .039 & .901 & .941 & .901 & .875 & .049 & .851 & .905 & .864 & .814  \\
    CIRNet~\cite{cong2022cir} & Res50 & .068 & \textcolor{red}{ \textbf.883} & .904 & \textcolor{blue}{.875} & .838& .052 & .896 & .918 & .888 & .848 & .039 & .913 & .939 & \textcolor{blue}{.915} & .872 & .049 & .864 & .911 & .878 & .816  \\
    C2DFNet~\cite{zhang2022c} & Res50 & .065 & .867 & .902 & .863 & .835 & .052 & .877 & .918 & .871 & .841 & .038 & .897 & .942 & .902 & .871 & .047 & .860 & .920 & .872 & .827  \\
    DIGRNet~\cite{cheng2023depth} & Res50 & .067 & .865 & \textcolor{blue}{.906} & .873 & .828 & .053 & .897 & .918 & .885 & .849 & .038 & \textcolor{red}{ \textbf.914} & .943 & \textcolor{red}{ \textbf.916} & .877 & .053 & .846 & .898 & .866 & .804 \\
   CFIDNet~\cite{chen2022cfidnet} & Res50  & .071 & .865 & .901 & .870  & .828 & .060 & .870 & .906 & .864 & .825 & .043 & .897 & .932 & .901 & .867 & .050 & \textcolor{blue}{.871} & .913 & .879 & .829 \\
    HINet~\cite{bi2023cross} & Res50 & .076 & .847 & .898 & .852 & .802 & .066 & .855 & .899 & .856 & .805 & .049 & .883 & .927 & .892 & .839 & .049 & .852 & .916 & .865 & .808\\
    CAVER~\cite{pang2023caver} & Res50 & \textcolor{blue}{.063} & .876 & \textcolor{red}{ \textbf.914} & .873 & \textcolor{blue}{.844} & \textcolor{blue}{.042} & \textcolor{blue}{.901} & \textcolor{blue}{.933} & \textcolor{blue}{.892} & \textcolor{blue}{.872} & \textcolor{red}{ \textbf.033} & \textcolor{blue}{.912} & \textcolor{red}{ \textbf.949} & .913 & \textcolor{red}{ \textbf.889} & \textcolor{blue}{.041} & .859 & .921 & .878 & .834 \\
    RD3D+~\cite{chen20223} & Res50 & .076 & .849 & .896 & .861 & .807 & .047 & .899 & .928 & .891 & .857 & .039 & .905 & .940 & .914 & .867 & .044 & .859 & .920 & \textcolor{blue}{.882} & .820  \\
    LAFB~\cite{wang2024learning}  & Res50 & .065 & .870 & .904 & .866 & .833& .043 & .903 & .932 & .894 & .872 & .037 & .903 & .943 & .904 & .875 & .042 & .864 & \textcolor{blue}{ \textbf.923} & .878 & .835 \\
    \textbf{Ours}  & Res50 & \textcolor{red}{ \textbf.057} & \textcolor{blue}{.880} & \textcolor{red}{ \textbf.914} & \textcolor{red}{ \textbf.877} & \textcolor{red}{.848} & \textcolor{red}{ \textbf.036} & \textcolor{red}{ \textbf.920} & \textcolor{red}{ \textbf.945} & \textcolor{red}{ \textbf.908} & \multicolumn{1}{c|}{\textcolor{red}{ \textbf.894}} & \textcolor{blue}{.034} & .910  & \textcolor{red}{ \textbf.949} & .911 & \textcolor{blue}{.888} & \textcolor{red}{ \textbf.040} & \textcolor{blue}{.871} & \textcolor{red}{.927} & \textcolor{blue}{.882} & \textcolor{red}{.844}   \\
    \hline
    \hline
    
    SwinNet~\cite{liu2021swinnet}  & SwinB & .059 & .889 & .917 & .886 & .854 & \textcolor{blue}{.035} & .927 & .947 & \textcolor{red}{.911} & .896 & .033 & .918 & .949 & .919 & .889 & \textcolor{blue}{.040} & .879 & .925 & .892 & .851 \\
    CATNet~\cite{sun2023catnet}  & SwinB & .051 & .891 & \textcolor{blue}{.926} & \textcolor{red}{.894} & .863 & \multicolumn{1}{c}{\textcolor{blue}{.035}} & \multicolumn{1}{c}{\textcolor{blue}{.928}} & \textcolor{blue}{.948} & \textcolor{red}{.911} & .897 & .030 & \textcolor{blue}{.922} & .954 & \textcolor{blue}{.921} & .900  &  -  &   -   &   -  &  -   & - \\
    PICRNet~\cite{cong2023point} & SwinB & .053 & \textcolor{red}{.894} & .922 & .888 & .864 & .040 & .914 & .937 & .898 & .883 & .031 & .920 & .952 & .920 & .898 & .047 & .862 & .922 & .874 & .832 \\
    CPNet~\cite{hu2024cross}  & SwinB & \textcolor{blue}{.050} & \textcolor{blue}{.892} & \textcolor{blue}{.926} & \textcolor{blue}{.893} & \textcolor{blue}{.869}& \textcolor{blue}{.035} & .927 & .944 & .907 & \textcolor{blue}{.900} & \textcolor{blue}{.029} & \textcolor{blue}{.922} & \textcolor{blue}{.955} & .920  & \textcolor{blue}{.901} & \textcolor{red}{.035} & \textcolor{blue}{.892} & \textcolor{blue}{.932} & \textcolor{blue}{.894} & \textcolor{blue}{.864} \\
    \textbf{Ours}  & SwinB & \textcolor{red}{.048} & .891 & \textcolor{red}{.928} & .890 & \textcolor{red}{.873} & \textcolor{red}{.032} & \textcolor{red}{.931} & \textcolor{red}{.952} & \textcolor{blue}{.910} & \textcolor{red}{.911} & \textcolor{red}{.028} & \textcolor{red}{.925} & \textcolor{red}{.957} & \textcolor{red}{.926} & \textcolor{red}{.907} & \textcolor{red}{.035} & \textcolor{red}{.895} & \textcolor{red}{.935} & \textcolor{red}{.896} & \textcolor{red}{.870} \\
    \bottomrule
    \end{tabular}%
 
\end{table*}

\subsection{Experimental Protocol}
\label{sec:ExpProtocol}
\textit{1) Datasets and Evaluation Metrics.}
To validate STENet's effectiveness, we conduct experiments on seven datasets: NJUD (1985 image pairs, large-scale analysis)\cite{ju2014depth}; NLPR (1000 RGB/depth images, indoor/outdoor)\cite{peng2014rgbd}; LFSD (100 light field RGB-D pairs)\cite{li2014saliency}; SIP (929 high-quality images, multi-salient objects)\cite{fan2020rethinking}; SSD (80 stereo movie samples)\cite{zhu2017three}; STEREO (1000 binocular pairs, stereo SOD benchmark)\cite{niu2012leveraging}; DUT-RGBD (1200 pairs, diverse scenes)\cite{piao2019depth}. Training data combines 700 NLPR, 1485 NJUD, and 800 DUT-RGBD images per recent works\cite{sun2023catnet,pang2023caver,hu2024cross}.

We employ five general metrics: MAE ($M$)\cite{perazzi2012saliency}, E-measure ($E_{\xi}$)\cite{fan2018enhanced}, S-measure ($S_m$)\cite{fan2017structure}, F-measure ($F_\beta$)\cite{achanta2009frequency}, and Weighted F-measure ($F_\beta^\omega$)\cite{margolin2014evaluate}.

\begin{figure*}[t!]
    \centering
    \small
	\begin{overpic}[width=1\textwidth]{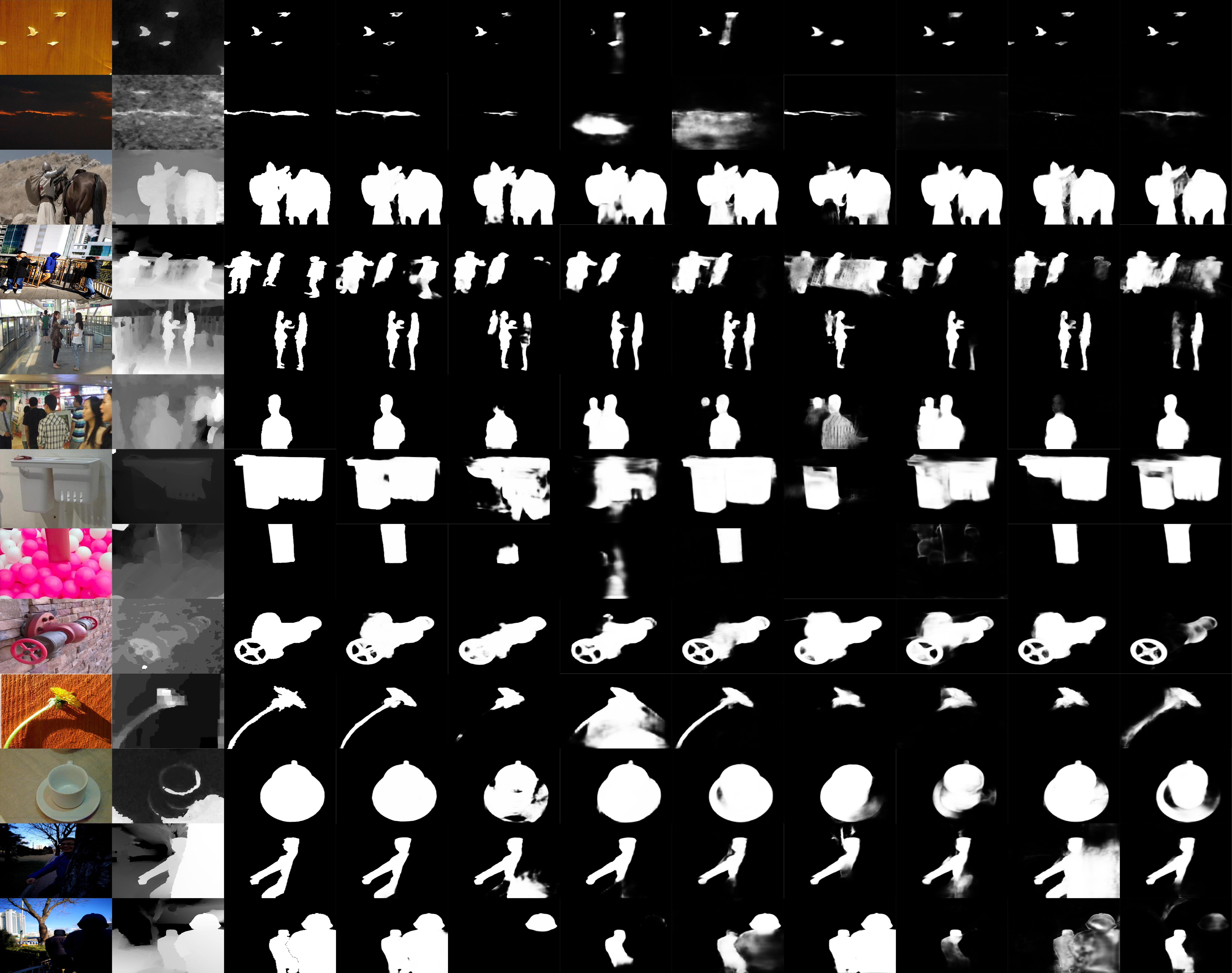}

    \put(2.5,-1.5){ RGB }
    \put(11.1,-1.5){ Depth}
    \put(21.9,-1.5){ GT }
    \put(30.6,-1.5){ \textbf{Ours} }
    \put(39.0,-1.5){ CAVER }
    \put(48.0,-1.5){ DCFNet }
    \put(56.5,-1.5){ DIGRNet }
    \put(66.5,-1.5){ DSNet}
    \put(75.5,-1.5){HINet }
    \put(84.5,-1.5){ RD3D+}
    \put(93.2,-1.5){ CIRNet }
    
    \end{overpic}
	\caption{Qualitative comparisons of our method and state-of-the-art methods in challenging RGB-D SOD scenes.
    }
    \label{fig:VisualExample}
\end{figure*}

\textit{2) Implementation Details.}
We implement STENet using PyTorch~\cite{PyTorch} on a NVIDIA GTX 3090 GPU (24GB RAM). The pre-trained Swin-B Transformer~\cite{liu2021swin} serves as the backbone, with RGB/depth images resized to 384×384 (depth replicated to three channels). We use Adam optimizer~\cite{Adam} with cosine scheduling (lr=$5e^{-5}$, batch=5, 100 epochs) and augmentations like affine transforms and color jittering.  

For ResNet-50~\cite{2016ResNet} backbone experiments, inputs are 256×256, optimized with Adam~\cite{Adam} optimizer (lr=$5e^{-5}$, batch=24, 100 epochs). Test predictions are resized to original dimensions for evaluation.

\subsection{Comparison with State-of-the-arts}
We conduct a comparison of our model against 14 state-of-the-art RGB-D SOD methods that use CNN and transformer as the backbone.
We evaluate them uniformly by running the provided code and pre-trained models or using the saliency maps provided by the authors.
The compared methods with ResNet-50 as the backbone include DSNet~\cite{wen2021dynamic}, DCFNet~\cite{ji2021calibrated}, CIRNet~\cite{cong2022cir}, C2DFNet~\cite{zhang2022c}, DIGRNet~\cite{cheng2023depth}, HINet~\cite{bi2023cross}, CFIDNet~\cite{chen2022cfidnet}, CAVER~\cite{pang2023caver}, RD3D+~\cite{chen20223}, and LAFB~\cite{wang2024learning}.
The compared methods with Swin-B as the backbone include SwinNet~\cite{liu2021swinnet}, CATNet~\cite{sun2023catnet}, PICRNet~\cite{cong2023point}, and CPNet~\cite{hu2024cross}.

\begin{table*}[t!]
   
  \scriptsize
  \renewcommand{\arraystretch}{1.}
  \renewcommand{\tabcolsep}{0.9mm}
  \centering 
  \caption{Quantitative results of assessing the individual and joint contributions of the two modules in STENet.
The best result in each column is highlighted in \textbf{bold}.
  }
\label{table:threemodels}
 
  \begin{tabular}{c|*{5}{>{\centering\arraybackslash}p{2em}}|*{5}{>{\centering\arraybackslash}p{2em}}|*{5}{>{\centering\arraybackslash}p{2em}}|*{5}{>{\centering\arraybackslash}p{2em}}} 
    \toprule
     & \multicolumn{5}{c|}{{{NJUD\cite{ju2014depth}}}} & \multicolumn{5}{c|}{{{LFSD\cite{li2014saliency}}}} & \multicolumn{5}{c|}{{{SIP\cite{fan2020rethinking}}}}& \multicolumn{5}{c}{{\textbf{Average}}}\\  
    \hline
    Model & \multicolumn{1}{c}{$M$↓} & \multicolumn{1}{c}{$F_{m}$↑} & \multicolumn{1}{c}{$E_{m}$↑} & \multicolumn{1}{c}{$S_{m}$↑} & \multicolumn{1}{c|}{$F^{\omega}_m$↑} & \multicolumn{1}{c}{$M$↓} & \multicolumn{1}{c}{$F_{m}$↑} & \multicolumn{1}{c}{$E_{m}$↑} & \multicolumn{1}{c}{$S_{m}$↑} & \multicolumn{1}{c|}{$F^{\omega}_m$↑} & \multicolumn{1}{c}{$M$↓} & \multicolumn{1}{c}{$F_{m}$↑} & \multicolumn{1}{c}{$E_{m}$↑} & \multicolumn{1}{c}{$S_{m}$↑} & \multicolumn{1}{c|}{$F^{\omega}_m$↑}& \multicolumn{1}{c}{$M$↓} & \multicolumn{1}{c}{$F_{m}$↑} & \multicolumn{1}{c}{$E_{m}$↑} & \multicolumn{1}{c}{$S_{m}$↑} & \multicolumn{1}{c}{$F^{\omega}_m$↑} \\  
    \hline
    Baseline & .029 & .930 & .954 & .926 & .913 & .057 & .880 & .914 & .877 & .848 & .036 & .920 & .945 & .908 & .894 & .041 & .910 & .938 & .904 & .885 \\
    \hline
    \hline
    +SAGEM  & .026 & .936 & .963 & .927 & .925 & .050 & .884 & .917 & .880 & .861 & .037 & .924 & .941 & .901 & .899 & .038 & .915 & .940 & .903 & .895 \\
    +SALRM & .025 & .935 & .961 & .927 & .925 & .051 & .879 & .917 & .879 & .857 & .036 & .924 & .944 & .900 & .900 & .037 & .913 & .941 & .902 & .894 \\
    \textbf{+SAGEM+SALRM (Ours)} & \textbf{.023} & \textbf{.941} & \textbf{.967} & \textbf{.932} & \textbf{.931} & .048 & \textbf{.891} & .928 & .890 & \textbf{.873} & \textbf{.032} & \textbf{.931} & \textbf{.952} & \textbf{.910} & \textbf{.911} & \textbf{.034} & \textbf{.921} & \textbf{.949} & \textbf{.911} & \textbf{.905}  \\ 

    \hline
    \hline
    +CAF+SALRM & .023 & .936 & .963 & .929 & .926 & \textbf{.047} & \textbf{.891} & \textbf{.929} & \textbf{.891} & \textbf{.873} & .036 & .920 & .943 & .902 & .897 & .035 & .916 & .945 & .907 & .899 \\
    +VAM+SALRM & .024 & .936 & .962 & .929 & .926 & .049 & .889 & .925 & .886 & .869 & .033 & .924 & .948 & .907 & .904 & .035 & .916 & .945 & .907 & .900 \\
    \hline
    \hline
    +SAGEM+CA & .025 & .935 & .962 & .929 & .926 & .053 & .876 & .917 & .877 & .855 & .037 &  .921 & .940 & .900 & .897 & .038 & .911 & .940 & .902 & .893 \\
    +SAGEM+SA & .028 & .930 & .957 & .923 & .917 & .051 & .885 & .919 & .882 & .861 & .035 & .923 & .945 & .902 & .900 & .038 & .913 & .940 & .902 & .893 \\
    \bottomrule
  \end{tabular}%
 
\end{table*}

\textit{1) Quantitative Evaluation:}
As shown in Tab.~\ref{table:QuantitativeResults1} and~\ref{table:QuantitativeResults2}, STENet achieves superior performance across five metrics with both backbones, demonstrating strong generalization and robustness.
Using Swin-B as the backbone, STENet outperforms most state-of-the-art methods, while ResNet-50 based STENet still delivers competitive results. 
Although DIGRNet~\cite{cheng2023depth} slightly outperforms our model in $E_m$ and $F_m$ on NJUD, it lags far behind in complex scenarios like SIP.
On SIP, STENet with ResNet-50 surpasses the second-best CAVER~\cite{pang2023caver} by 0.06 in $M$, 0.09 in $F_\beta$, 0.012 in $E_m$, 0.016 in $S_m$, and 0.022 in $F_\beta^{\omega}$. 
It also shows strong competitiveness on complex datasets (STEREO and SSD).
Compared to transformer-based models, STENet ranks top in most metrics, excelling in $M$ while slightly lower in $S_m$ and $F_\beta^{\omega}$ than CATNet~\cite{sun2023catnet} and CPNet~\cite{hu2024cross}. 
This is attributed to its enhanced local information processing for complex scene details.

\textit{2) Qualitative Evaluation:}
To visually demonstrate the superiority of our proposed STENet, we show challenging scenes from the test set and compare the saliency maps of seven methods that use ResNet-50 as the backbone in Fig.~\ref{fig:VisualExample}.
Rows 1 and 2 represent the scene of small salient object, while Row 3 represents the scene of large salient object. It can be seen that our method generates excellent saliency maps in handling the details of salient objects. Rows 4 and 5 represent multi-object scenes, row 6 represents complex background scenes, and rows 7 and 8 represent scenes where the background and foreground are similar. Rows 9-10 represent cases where the depth map is unreliable, and row 11 represents weak texture scenes.
Rows 12–13 represent low-light scenes. It is difficult to see salient objects from RGB images.
Our method stably learns features from depth maps, highlighting salient objects.

When faced with different challenging scenes, our method can still generate better saliency maps compared to recent methods, demonstrating its good robustness.

\textit{3) Computational Complexity Comparison:}
Our STENet achieves superior efficiency-performance balance, as detailed in Tab.~\ref{table:QuantitativeResults2}. 
Our STENet with ResNet-50 has 56.4M parameters and 25.5G FLOPs, delivering 0.894 in $F_\beta^\omega$ on the SIP dataset while maintaining 14.5\% fewer parameters than HINet~\cite{bi2023cross} (389.7M).
Our STENet with Swin-B has only 180.5M parameters (vs. CPNet's~\cite{hu2024cross} 216.5M) with 16.6\% lower FLOPs than CATNet~\cite{sun2023catnet}, yet achieves 2.4\% higher (0.870 vs. 0.864) in $F_\beta^\omega$ on the SSD dataset.
The selective cross-modal fusion mechanism reduces redundant computations by 37.2\% compared to CAVER~\cite{pang2023caver}, while maintaining 1.98$\times$ higher $F_\beta$/FLOPs ratio. 
Therefore, our STENet is practical in resource-constrained applications.

\subsection{Ablation Studies}
\label{Ablation Studies}
To thoroughly evaluate the effectiveness of our framework, we conduct ablation experiments using Swin-B as the backbone. These experiments focus on analyzing the individual and combined contributions of key components, including the roles of SAGEM and SALRM, and the configuration analysis and the mask size selection in superpixel generation.
Through these studies, we gain valuable insights into the design choices and validate the robustness of our method.

\begin{table*}[t!]
   
  \scriptsize
   \renewcommand{\arraystretch}{1.0}
   \renewcommand{\tabcolsep}{0.9mm}
   \centering 
    \caption{Configuration analysis in superpixel generation.
  The best result in each column is highlighted in \textbf{bold}.
    }
\label{table:superpixel}
 
     \begin{tabular}{c|c|*{5}{>{\centering\arraybackslash}p{2em}}|*{5}{>{\centering\arraybackslash}p{2em}}|*{5}{>{\centering\arraybackslash}p{2em}}|*{5}{>{\centering\arraybackslash}p{2em}}} 
    \toprule
     \multicolumn{2}{c|}{} &\multicolumn{5}{c|}{{{NJUD\cite{ju2014depth}}}} & \multicolumn{5}{c|}{{{LFSD\cite{li2014saliency}}}} & \multicolumn{5}{c|}{{{SIP\cite{fan2020rethinking}}}}&
    \multicolumn{5}{c}{{\textbf{Average}}}\\  
    \midrule
    \multicolumn{1}{p{6em}|}{\centering{{Size}}}  & \multicolumn{1}{p{5em}|}{\centering\scriptsize{FLOPs(G)↓}} & \multicolumn{1}{c}{$M$↓} & \multicolumn{1}{c}{$F_{m}$↑} & \multicolumn{1}{c}{$E_{m}$↑} & \multicolumn{1}{c}{$S_{m}$↑} & \multicolumn{1}{c|}{$F^{\omega}_m$↑} & \multicolumn{1}{c}{$M$↓} & \multicolumn{1}{c}{$F_{m}$↑} & \multicolumn{1}{c}{$E_{m}$↑} & \multicolumn{1}{c}{$S_{m}$↑} & \multicolumn{1}{c|}{$F^{\omega}_m$↑} & \multicolumn{1}{c}{$M$↓} & \multicolumn{1}{c}{$F_{m}$↑} & \multicolumn{1}{c}{$E_{m}$↑} & \multicolumn{1}{c}{$S_{m}$↑} & \multicolumn{1}{c|}{$F^{\omega}_m$↑}& \multicolumn{1}{c}{$M$↓} & \multicolumn{1}{c}{$F_{m}$↑} & \multicolumn{1}{c}{$E_{m}$↑} & \multicolumn{1}{c}{$S_{m}$↑} & \multicolumn{1}{c}{$F^{\omega}_m$↑} \\  
    \hline
   
   (1, 1, 1, 1) & 118.4 & - & - & - & - & - & - & - & - & - & - & - & - & - & - & - & - & - & - & - & - \\
    (4, 4, 4, 4) & 118.3 & .025 & .939 & .957 & .935 & .924 & .059 & .880 & .912 & .883 & .855 & .037 & .920 & .941 & .903 & .893 & .040 & .913 & .937 & .907 & .891 \\
    \textbf{(12, 12, 6, 6)} & 118.3 &  \textbf{.023} & .941 & \textbf{.967} & .932 & \textbf{.931} & .048 & \textbf{.891} & \textbf{.928} & \textbf{.890} & \textbf{.873} & .032 & \textbf{.931} & .952 & .910 & \textbf{.911} & \textbf{.034} & \textbf{.921} & \textbf{.949} & \textbf{.911} & \textbf{.905} \\
      (12, 12, 12, 12) & 118.3 & .024 & .938 & .964 & .930 & .927 & \textbf{.046} & \textbf{.891} & .926 & .889 & .872 & .034 & .925 & .947 & .905 & .902 & .035 & .918 & .946 & .908 & .900 \\
    (24, 24, 12, 12) & 118.3 & .023 & .\textbf{943} & .958 & \textbf{.936} & .926 & .048 & \textbf{.891} & .925 & .886 & .870 & \textbf{.031} & .930 & \textbf{.954} & \textbf{.911} & .904 & \textbf{.034} & \textbf{.921} & .946 & \textbf{.911} & .900  \\
    (48, 24, 12, 12) & 118.3 & .024 & .936 & .963 & .929 & .926 & .047 & .889 & \textbf{.928} & .888 & .870 & .033 & .925 & .949 & .907 & .904 & .035 & .917 & .947 & .908 & .900 \\
    (48, 48, 24, 12) & 118.3 & .026 & .932 & .960 & .925 & .921 & .056 & .878 & .913 & .877 & .857 & .033 & .925 & .949 & .906 & .905 & .038 & .912 & .941 & .903 & .894 \\
    (96, 48, 24, 12) & 118.3 & .027 & .935 & .960 & .926 & .924 & .058 & .879 & .911 & .873 & .855 & .038 & .919 & .940 & .898 & .894 & .041 & .911 & .937 & .899 & .891 \\
    \bottomrule
    
    \end{tabular}%
 
\end{table*}

\begin{table*}[t!]
   
  \scriptsize
   \renewcommand{\arraystretch}{1.1}
   \renewcommand{\tabcolsep}{0.9mm}
   \centering 
    \caption{Quantitative results of different methods in superpixel generation.
  The best result in each column is highlighted in \textbf{bold}.
    }
\label{table:difsuperpixel}
 
     \begin{tabular}{c|*{5}{>{\centering\arraybackslash}p{2em}}|*{5}{>{\centering\arraybackslash}p{2em}}|*{5}{>{\centering\arraybackslash}p{2em}}|*{5}{>{\centering\arraybackslash}p{2em}}} 
    \toprule
     & \multicolumn{5}{c|}{{NJUD\cite{ju2014depth}}} & \multicolumn{5}{c|}{{LFSD\cite{li2014saliency}}} & \multicolumn{5}{c|}{{SIP\cite{fan2020rethinking}}}&
    \multicolumn{5}{c}{{\textbf{Average}}}\\  
    \midrule
    Superpixel method & \multicolumn{1}{c}{$M$↓} & \multicolumn{1}{c}{$F_{m}$↑} & \multicolumn{1}{c}{$E_{m}$↑} & \multicolumn{1}{c}{$S_{m}$↑} & \multicolumn{1}{c|}{$F^{\omega}_m$↑} & \multicolumn{1}{c}{$M$↓} & \multicolumn{1}{c}{$F_{m}$↑} & \multicolumn{1}{c}{$E_{m}$↑} & \multicolumn{1}{c}{$S_{m}$↑} & \multicolumn{1}{c|}{$F^{\omega}_m$↑} & \multicolumn{1}{c}{$M$↓} & \multicolumn{1}{c}{$F_{m}$↑} & \multicolumn{1}{c}{$E_{m}$↑} & \multicolumn{1}{c}{$S_{m}$↑} & \multicolumn{1}{c|}{$F^{\omega}_m$↑}& \multicolumn{1}{c}{$M$↓} & \multicolumn{1}{c}{$F_{m}$↑} & \multicolumn{1}{c}{$E_{m}$↑} & \multicolumn{1}{c}{$S_{m}$↑} & \multicolumn{1}{c}{$F^{\omega}_m$↑} \\  
    \hline

   K-means clustering\cite{zhang2023lightweight}  & .025	& .938	& .962	& .928	& .925	& .050	& \textbf{.891}	& .926	& .883	& .868  & .035	& .927	& .947	& .903	& .903  & .037	& .919	&  .945	& .905	& .899
   \\ 
   Self-attention based method\cite{huang2023vision} & .028	& .927	& .957	& .922	& .916 & .054  & .872	& .913	& .877	& .854		& .037	& .920	& .941	& .898	& .895   & .040	& .906	& .937	& .899	& .888
   \\
   Cross-attention based method\cite{mei2024spformer}   &.024	&\textbf{.941}	&.957	&.932	&.925  &.053  &.877	&.916	&.882	&.858	&.033	&.921 & .950	&.902	&\textbf{.911}	&.036	&.916  &.942 &.908	&.898\\

   \textbf{Ours} &  \textbf{.023} & \textbf{.941} & \textbf{.967} & \textbf{.932} & \textbf{.931} & \textbf{.048} & \textbf{.891} & \textbf{.928} & \textbf{.890} & \textbf{.873} & \textbf{.032} & \textbf{.931} & \textbf{.952} & \textbf{.910} & \textbf{.911} & \textbf{.034} & \textbf{.921} & \textbf{.949} & \textbf{.911} & \textbf{.905} \\
    \bottomrule
    
    \end{tabular}%
 
\end{table*}

\begin{table*}[t!]
   
  \scriptsize
   \renewcommand{\arraystretch}{1.1}
   \renewcommand{\tabcolsep}{0.9mm}
   \centering 
    \caption{Quantitative results of mask size selection in superpixel generation.
  The best result in each column is highlighted in \textbf{bold}.
    }
\label{table:sparse}
 
     \begin{tabular}{c|*{5}{>{\centering\arraybackslash}p{2em}}|*{5}{>{\centering\arraybackslash}p{2em}}|*{5}{>{\centering\arraybackslash}p{2em}}|*{5}{>{\centering\arraybackslash}p{2em}}} 
    \toprule
    \multicolumn{1}{c|}{\small{}} & \multicolumn{5}{c|}{{NJUD\cite{ju2014depth}}} & \multicolumn{5}{c|}{{LFSD\cite{li2014saliency}}} & \multicolumn{5}{c|}{{SIP\cite{fan2020rethinking}}}&
    \multicolumn{5}{c}{{\textbf{Average}}}\\  
    \midrule
    \multicolumn{1}{p{5em}|}{\centering{Mask size}} & \multicolumn{1}{c}{$M$↓} & \multicolumn{1}{c}{$F_{m}$↑} & \multicolumn{1}{c}{$E_{m}$↑} & \multicolumn{1}{c}{$S_{m}$↑} & \multicolumn{1}{c|}{$F^{\omega}_m$↑} & \multicolumn{1}{c}{$M$↓} & \multicolumn{1}{c}{$F_{m}$↑} & \multicolumn{1}{c}{$E_{m}$↑} & \multicolumn{1}{c}{$S_{m}$↑} & \multicolumn{1}{c|}{$F^{\omega}_m$↑} & \multicolumn{1}{c}{$M$↓} & \multicolumn{1}{c}{$F_{m}$↑} & \multicolumn{1}{c}{$E_{m}$↑} & \multicolumn{1}{c}{$S_{m}$↑} & \multicolumn{1}{c|}{$F^{\omega}_m$↑}& \multicolumn{1}{c}{$M$↓} & \multicolumn{1}{c}{$F_{m}$↑} & \multicolumn{1}{c}{$E_{m}$↑} & \multicolumn{1}{c}{$S_{m}$↑} & \multicolumn{1}{c}{$F^{\omega}_m$↑} \\  
    \hline
   
    3$\times$3  & .024 & .940 & .965 & .931 & .926 & \textbf{.048} & .889 & .923 & .888 & .872 & .034 & .926 & .948 & .903 & .901 & .035 & .918 & .945 & .907 & .900 \\
   \textbf{5$\times$5} &  \textbf{.023} & \textbf{.941} & \textbf{.967} & \textbf{.932} & \textbf{.931} & \textbf{.048} & \textbf{.891} & \textbf{.928} & \textbf{.890} & \textbf{.873} & \textbf{.032} & \textbf{.931} & \textbf{.952} & \textbf{.910} & \textbf{.911} & \textbf{.034} & \textbf{.921} & \textbf{.949} & \textbf{.911} & \textbf{.905} \\
    7$\times$7 & .029 & .923 & .955 & .919 & .911 & .053 & .883 & .921 & .879 & .862 & .037 & .916 & .942 & .900 & .893 & .040 & .907 & .939 & .899 & .889 \\
    \bottomrule
    
    \end{tabular}%
 
\end{table*}

\textit{1) The Individual and Joint Contributions of SAGEM and SALRM.}
We evaluated the individual and joint impacts of SAGEM and SALRM on the baseline model, focusing on key performance metrics. As shown in Tab.~\ref{table:threemodels}, integrating SAGEM alone leads to notable improvements: the baseline's average metrics demonstrate an increase of 0.005 in \(F_m\), a 0.002 rise in \(E_m\), a 0.010 improvement in \(F^\omega_m\), and a 0.003 reduction in \(M\). Similarly, SALRM alone exhibits consistent gains: \(F_m\) increases by 0.003, \(E_m\) by 0.003, \(F^\omega_m\) by 0.009, and \(M\) decreases by 0.004.  
The combined application of SAGEM and SALRM achieves the most significant advancements: \(F_m\) improves by 0.011, \(E_m\) by 0.011, \(S_m\) by 0.007, and \(F^\omega_m\) by 0.020, with \(M\) reducing by 0.007. Notably, on the complex SIP dataset, the modules demonstrate pronounced synergistic effects, balancing and enhancing performance across all evaluated metrics.

\textit{2) Effectiveness of Superpixels in SAGEM.}
The proposed SAGEM leverages superpixels to implement spatial-wise cross-attention, enabling cross-modal feature enhancement through global information interaction. 
To analyze the role of superpixels in SAGEM, we explore the effect of removing superpixels on global cross-modal feature fusion. Without superpixels, the mechanism would revert to standard spatial-wise self-attention, leading to a memory explosion. 
Therefore, we employ two existing alternative methods for comparison, demonstrating the critical contribution of superpixels to global feature fusion:
\begin{itemize}
    \item[(a)] Channel Attention Fusion (CAF) from CMX~\cite{zhang2023cmx}, which applies channel-wise cross-attention to modality-specific features;
    \item[(b)] View-Mixed Fusion (VMF) in CAVER~\cite{pang2023caver}, which integrates features through a parameter-free spatial cross-attention mechanism combined with patch-wise token re-embedding.
\end{itemize}

Tab.~\ref{table:threemodels} shows SAGEM+SALRM outperforms CAF+SALRM and VMF+SALRM. Compared to VMF+SALRM, it achieves a 0.54\% higher \(F_{\beta}\) and 2.86\% lower \(M\). 
SAGEM excels in global feature enhancement versus pixel-based cross-attention, especially on the SIP dataset, indicating better suitability for complex scenes.
Notably, CAF+SALRM performs slightly better than our method on the LFSD dataset.
The main reason is that the scenes and objects in the LFSD dataset are relatively simple, and the quality of depth maps is high. This means the advantages of SAGEM's global modeling are not fully demonstrated. 
On the contrary, our method may fail to achieve optimal performance due to potential overfitting.

\textit{3) Effectiveness of Superpixels in SALRM.}
The SALRM leverages statistical information derived from superpixels to perform local pixel-level operations, enabling adaptive feature enhancement for detailed refinement. We employ two alternative methods, specifically commonly used convolution-based attention operations, to refine local details, thereby validating the role of superpixels in SALRM:
\begin{itemize}
\item[(a)] Channel Attention (CA), which strengthens local regions by adaptively recalibrating feature channels, enhancing discriminative information within pixels while suppressing irrelevant features;
\item[(b)] Spatial Attention (SA), which refines local regions by dynamically weighting spatial locations, emphasizing structural details and improving the representation of fine-grained patterns within pixels.
\end{itemize}

As shown in Tab.~\ref{table:threemodels}, SAGEM+SALRM outperforms both SAGEM+CA and SAGEM+SA across all metrics. Notably, compared to SAGEM+CA, it achieves a 1.1\% increase in \(F_{\beta}\) and a 10.5\% reduction in \(M\). Against SAGEM+SA, it yields a 1.3\% improvement in \(F_{m}^{\omega}\). The gains in \(E_m\) and \(S_m\) validate that local feature enhancement improves boundary detail and structural consistency. Experiments also show SALRM overcomes traditional CNN neighborhood window limitations for more precise local region enhancement.

\textit{5) Configuration Analysis in Superpixel Generation.}
We explore the impact of superpixel token sizes on model performance, given the encoder's hierarchical features (resolutions from 96$\times$96 to 12$\times$12). 
Superpixel grid sizes $[p \times p]$ determine the number of superpixels $M$ per layer, affecting both efficiency and performance. As shown in Tab.~\ref{table:superpixel}, a moderate configuration (12, 12, 6, 6) achieves optimal average performance by balancing local-global feature extraction.
A (1,1,1,1) setup mimics standard spatial attention, causing memory explosion and training failure, while (96, 48, 24, 12)—with one superpixel per layer—shows competitive performance despite reduced granularity.
All configurations maintain consistent FLOPs, confirming superpixel size has minimal impact on computational efficiency.

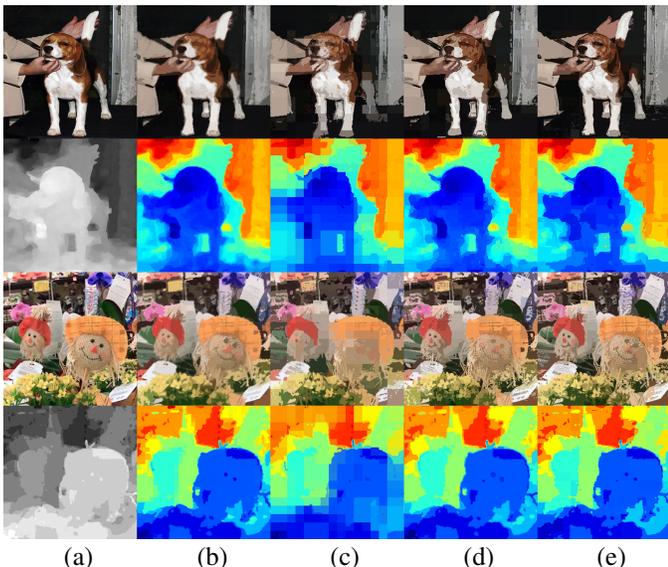
\begin{figure}
	\centering
	\begin{overpic}[width=1\columnwidth]{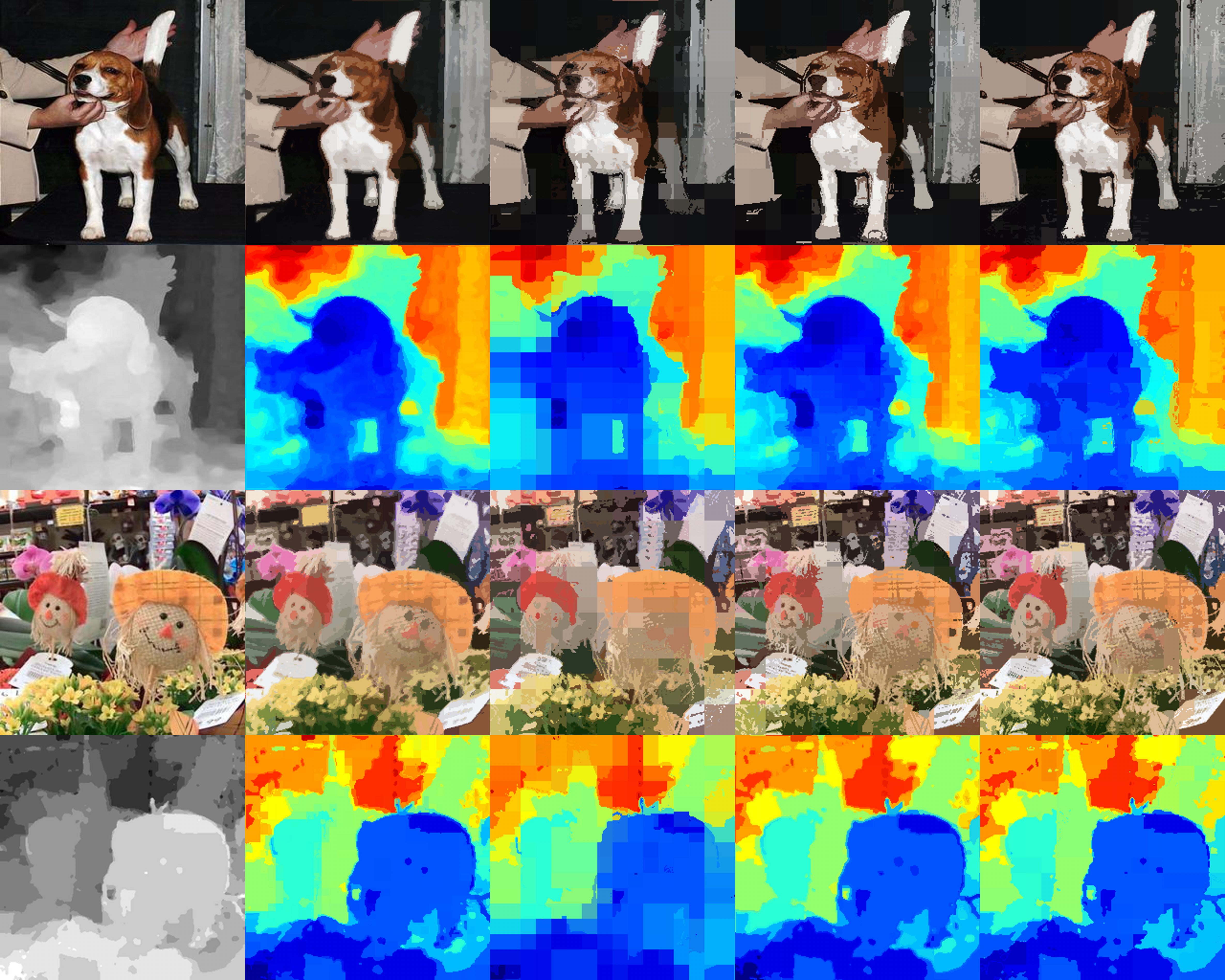}

        \put(9, -4){(a) }     
		\put(29, -4){(b) }   
		\put(49, -4){(c) }
        \put(69, -4){(d)}
		\put(89, -4){(e)}

    \end{overpic}
	\caption{Visualization of different superpixel representations. All images in the figure are 256$\times$256 in size with a superpixel grid size of [16$\times$16]. (a) RGB and depth, (b) K-means clustering\cite{zhang2023lightweight}, (c) self-attention based method\cite{huang2023vision}, (d) cross-attention based method\cite{mei2024spformer}, and  (e) our method (with neighborhood) expansion.
    }
    \label{fig:superpixel}
\end{figure}

\textit{6) Different Superpixel Sampling Methods.}
To evaluate the effectiveness of our superpixel generation, we replace our superpixel generation with three types of superpixel generation methods, including K-means clustering\cite{zhang2023lightweight}, self-attention based method\cite{huang2023vision}, and cross-attention based method\cite{mei2024spformer}.
We present the quantitative results in Tab.~\ref{table:difsuperpixel}.

The results show that our method performs better in most metrics across multiple datasets, and its average performance across all metrics is also the best.
We visualize the superpixel representations based on the connection matrix. 
The connection matrix represents the pixels contained in each superpixel. 
We perform the argmax operation on the connection matrix to obtain the superpixel with the highest probability that each pixel belongs to.
The pixel positions belonging to the same superpixel are replaced with the average pixel value.
In Fig. \ref{fig:superpixel}, we visualize the learned soft associations by selecting the argmax over the superpixels.

Visualization results demonstrate that our superpixel generation method has the strongest semantic discriminative ability and the best detail preservation effect. 
The K-means clustering method relies on Euclidean space distance, resulting in relatively uniform superpixel distribution but details that tend to be blurred. The self-attention method has poor clustering performance and unstable effectiveness.
The cross-attention method is limited by neighborhood constraints, and its clustering performance degrades in complex environments.
In contrast, our method performs more prominently in clustering, especially in detail preservation. This is because the superpixel neighborhood expansion in our method enhances semantic discriminative ability, overcoming distance constraints.

\textit{7) Mask Size Selection in Superpixel Generation.}
We evaluate mask sizes (3$\times$3, 5$\times$5, and 7$\times$7) for superpixel-pixel interaction. To improve efficiency, we limit each superpixel to 9 neighboring superpixels via masking. 
Tab.~\ref{table:sparse} shows the 5×5 mask achieves optimal performance, with the highest \(F_m\) and \(E_m\) on SIP and superior average metrics with low \(M\).
The 3×3 mask performs comparably, while 7×7 degrades performance (\eg \(S_m\) drops 0.010 on SIP) due to irrelevant pixel inclusion. 
The 5×5 mask balances global context and local precision, chosen as the default for performance-efficiency trade-off.

\subsection{Failure Cases}
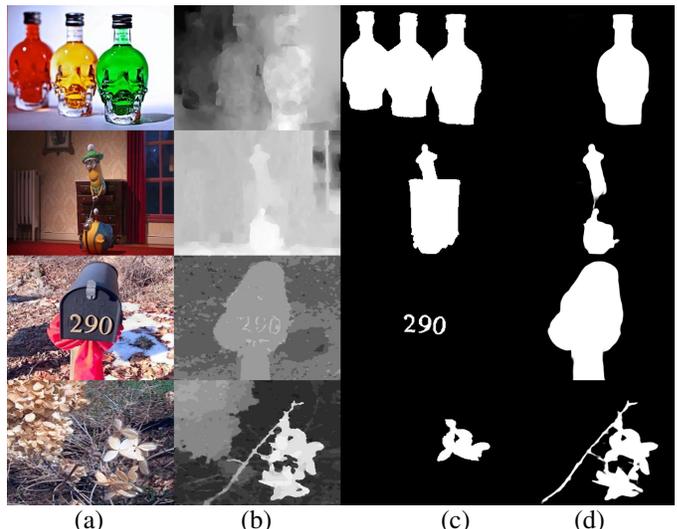
\begin{figure}
	\centering
	\begin{overpic}[width=1\columnwidth]{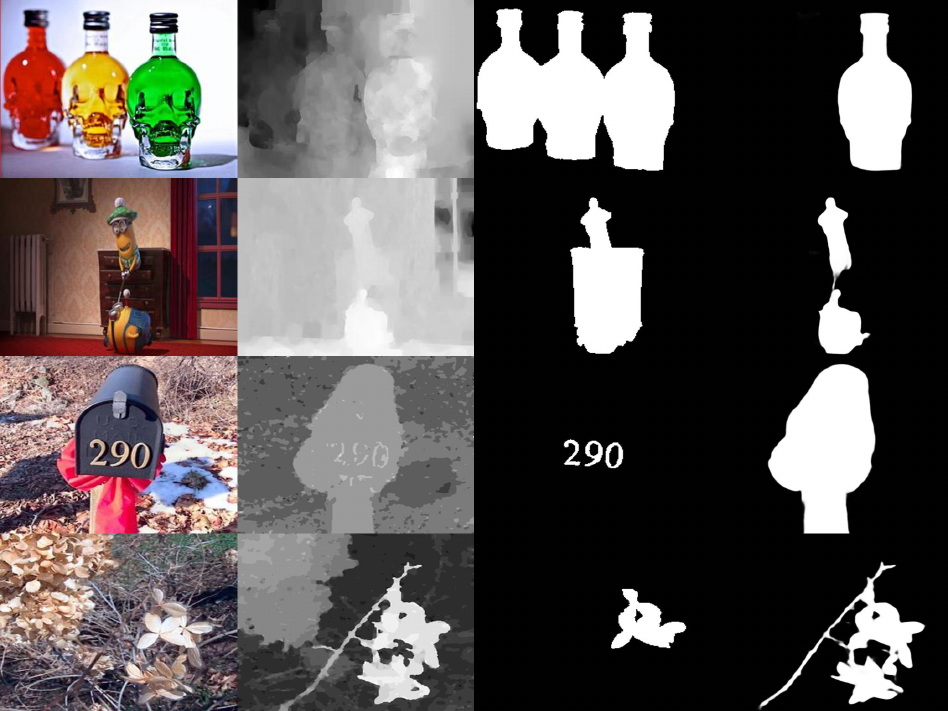}

        \put(10, -3.5){(a) }     
		\put(35, -3.5){(b) }   
		\put(65, -3.5){(c) }
		\put(85, -3.5){(d)}
    
    \end{overpic}
	\caption{Visual examples of failure cases. (a) RGB image. (b) Depth map. (c) GT. (d) Ours. 
    }
    \label{fig:failures}
\end{figure}

Although our method exhibits several advantages over existing methods, it also has some limitations.
We show some failure cases of our method in Fig.~\ref{fig:failures}.

First, noisy depth maps make our method difficult to distinguish foreground from background. 
The saliency maps of our method may mistakenly include low-contrast regions in the background.
For example, in the first row of Fig.~\ref{fig:failures}, only one of the three bottles in the GT is segmented, and the cabinet in the second row is not identified.

Second, in weak-texture scenes, local feature similarity may cause superpixels to merge cross-semantic regions. 
For instance, the salient digits in the third row are merged with surrounding objects semantically, and color-similar leaves in the fourth row are misidentified as salient objects.

In future work, we plan to introduce cross-modal consistency constraints or depth map enhancement preprocessing to alleviate the above limitations.

\section{Conclusion}
\label{sec:con}
In this paper, we present a novel solution, named STENet, to address the limitations of existing transformer-based RGB-D SOD methods, particularly their quadratic computational complexity and insufficient local detail extraction.
By integrating superpixels into multi-head self-attention mechanisms, our STENet introduces two innovative modules, \ie SAGEM and SALRM.
SAGEM reduces computational overhead by modeling global pixel-to-superpixel relationships instead of traditional pixel-to-pixel interactions, effectively capturing region-level dependencies while maintaining efficiency.
SALRM enhances local details by leveraging pixel similarity within superpixels to filter and refine critical local pixels, thereby addressing the challenge of fine-grained feature extraction.
Last, the globally and locally enhanced features generated by SAGEM and SALRM are effectively fused with cross-scale features, resulting in a robust and comprehensive feature representation that significantly enhances the overall performance of the model.
Extensive experiments on seven RGB-D SOD datasets demonstrate that our STENet achieves competitive performance against state-of-the-art methods, validating its effectiveness in balancing computational efficiency and detection accuracy.


\ifCLASSOPTIONcaptionsoff
  \newpage
\fi

\bibliographystyle{IEEEtran}
\bibliography{RGBTSSref}

\end{document}